\pdfoutput=1

\documentclass[11pt]{article}
\usepackage[preprint]{acl}
\usepackage{times}
\usepackage{latexsym}
\usepackage{footnote}
\usepackage{booktabs}
\usepackage{multirow} 
\usepackage{graphicx}
\usepackage{amssymb}
\usepackage{bbding}
\usepackage{makecell}
\usepackage[T1]{fontenc}

\usepackage[utf8]{inputenc}

\usepackage{microtype}

\usepackage{inconsolata}

\usepackage{graphicx}

\title{ReCoQA: A Benchmark for Tool-Augmented and Multi-Step Reasoning in Real Estate Question and Answering}

\author{ \textbf{Yindong Zhang\textsuperscript{1,2}},
 \textbf{Wenmian Yang\textsuperscript{3}\thanks{Corresponding author
}},
 \textbf{Yiquan Zhang\textsuperscript{3}},
 \textbf{Weijia Jia\textsuperscript{2,3}}
\\
\textsuperscript{1}Hong Kong Baptist University
\textsuperscript{2}Beijing Normal-Hong Kong Baptist University\\
\textsuperscript{3}Beijing Normal University\\
\texttt{yindongzhang@uic.edu.cn, \{wenmianyang, jiawj\}@bnu.edu.cn}, \\
\texttt{zhangyq987@hotmail.com}
}

\begin{document}
\maketitle
\begin{abstract}
Developing agents capable of navigating fragmented, multi-source information remains challenging, primarily due to the scarcity of benchmarks reflecting hybrid workflows combining database querying with external APIs. To bridge this gap, we introduce ReCoQA, a large-scale benchmark of 29,270 real-estate instances featuring machine-verifiable supervision for intermediate steps, including structured intent labels, SQL queries, and API calls. Complementarily, we propose HIRE-Agent, a hierarchical framework instantiating an understand–plan–execute architecture as a strong baseline. By orchestrating a Front-end parser, a planning Supervisor, and execution Specialists, HIRE-Agent effectively integrates heterogeneous evidence. Extensive experiments demonstrate that HIRE-Agent constitutes a strong baseline and substantiates the necessity of hierarchical collaboration for complex, real-world reasoning tasks. The benchmark and source code are available at: \url{https://github.com/Husky-989/ReCoQA}

\end{abstract}

\section{Introduction}
Real estate decision-making is a cognitively demanding process that requires synthesizing vast amounts of heterogeneous information. For prospective homebuyers, this workflow is often fragmented and inefficient. Users are compelled to navigate disparate platforms: comparing listings on one site, computing commute times on map applications, and verifying school districts or infrastructure plans on government portals. This fragmentation creates a substantial information barrier, leading to high time costs and cognitive overload.

While human real estate agents help bridge this gap, their services are labor-intensive, difficult to scale, and prone to interest-driven bias. Compared with traditional dialogue system \cite{qin2021co}, autonomous AI agents present a promising alternative for providing objective, efficient, and accessible consultation \cite{xie2025awareness,mou2024individual}. However, developing such agents requires capabilities far beyond traditional Question Answering (QA) systems. Real-world queries in this domain are inherently exploratory and compositional, often involving implicit constraints that require hybrid reasoning over decoupled data sources. Consider the query: ``Find secondhand apartment complexes in Tianhe District with a commute to Jimanjia Plaza under 30 minutes, within a 2km walk to Junjing Primary School, and priced between 30k-40k CNY/sqm.'' Answering this request necessitates a sophisticated interplay of semantic understanding, multi-step reasoning, and the orchestration of diverse tools (e.g., SQL databases and map APIs).

Building an intelligent system capable of handling such complexity entails overcoming three interconnected technical challenges. First, precise intent parsing is required to map colloquial and implicit expressions to machine-executable parameters (e.g., inferring transport modes from ``commute under 30 minutes''). Second, autonomous planning is essential to decompose high-level user goals into executable reasoning chains, determining the optimal sequence of sub-tasks. Third, heterogeneous integration is necessary to synthesize evidence from diverse formats, e.g., tabular records from relational databases and JSON objects from APIs, to derive a faithful final answer.


Despite these needs, systematic progress has been hindered by the scarcity of benchmarks that accurately reflect this complexity. While classical datasets like Spider \cite{yu2018spider} focus on structured querying, and recent agent benchmarks \cite{zhu2025multiagentbench, xu2025crab} evaluate general tool usage, they rarely emulate the strict interdependence required in vertical domains. Most existing benchmarks treat database querying and external API calls as isolated capabilities. They fail to capture hybrid workflows where the output of a database query (e.g., a list of candidate communities) must dynamically serve as the input for external verification tools (e.g., distance calculation APIs). This gap prevents the rigorous evaluation of agents in realistic, multi-source environments.

To address this, we introduce Real Estate Complex Question Answering (ReCoQA), a benchmark designed to evaluate an agent's end-to-end ability to understand, plan, and execute multi-step reasoning tasks. We contribute to the field on two fronts: data and methodology.

On the data front, we release the ReCoQA dataset, comprising 29,270 question-answer pairs grounded in authentic home-buying scenarios across eight major Chinese cities. The dataset covers residential communities and Points of Interest (POIs) and features three progressively challenging query types: single-step table QA, joint Database-API queries, and composite queries requiring reasoning beyond tool use. Crucially, to enable interpretability and fine-grained supervision, each sample is annotated with a machine-verifiable chain of intermediate steps, including structured Spoken Language Understanding (SLU) labels, SQL queries, and API call sequences.

Given the hybrid nature of ReCoQA, which couples structured database querying with external API interactions, there exists no directly comparable end-to-end baseline in prior work. To establish a rigorous evaluation standard, we implement the Hierarchical Intelligent Real Estate Agent (HIRE-Agent) as a benchmark and diagnostic system. 
Rather than proposing a novel architecture, HIRE-Agent serves as a strong baseline for examining the necessity of structured decomposition in this domain. Adopting an ``Understand--Plan--Execute'' cognitive flow, the framework coordinates a Front-end Agent for intent parsing, a Supervisor Agent for task orchestration, and Specialist Agents for tool execution. This hierarchical design decouples schema linking and API routing, enabling systematic evaluation of heterogeneous evidence integration in ReCoQA.



Our main contributions are summarized as follows:
\begin{itemize}
\item We define the ReCoQA task and release the first large-scale benchmark specifically designed to foster research on hybrid, multi-source reasoning in realistic decision-making scenarios.
\item We develop HIRE-Agent as a strong baseline system. By benchmarking against this hierarchical framework, we empirically validate that effective collaboration between planning and specialist agents is a prerequisite for solving complex, real-world queries.

\item We conduct extensive experiments on ReCoQA, providing a detailed breakdown of current LLM capabilities and identifying critical bottlenecks to guide future research in vertical domain agents.
\end{itemize}

\begin{table*}[htbp]
\centering
\scalebox{0.95}{
\begin{tabular}{ccccccc}
\toprule
Dataset & QA Pairs & SLU & Tool Calls & Multi-resources & Reasoning & Intermediate Result\\
\toprule
TableBench & 886 & \XSolidBrush & \XSolidBrush & \XSolidBrush & \Checkmark & \XSolidBrush\\
CRT-QA & 1,000 & \XSolidBrush & \XSolidBrush &\XSolidBrush & \Checkmark & \Checkmark \\
MapQA & 3,154 & \XSolidBrush & \XSolidBrush & \XSolidBrush & \Checkmark &\XSolidBrush\\
RETQA & 20,762 & \Checkmark & \XSolidBrush & \XSolidBrush & \XSolidBrush &\Checkmark\\
ReCoQA (Ours) & 29,270 & \Checkmark & \Checkmark & \Checkmark & \Checkmark& \Checkmark\\
\bottomrule
\end{tabular}}
\caption{Dataset Comparison.}
\label{tab:cmp}
\end{table*}

\section{Related Work}
\subsection{Complex TQA datasets}
Tabular Question Answering (TQA) datasets have evolved from early closed-domain tasks \cite{pasupat2015compositional, zhong2017seq2sql, yu2018spider} to complex reasoning benchmarks. Recent work has focused on numerical reasoning \cite{chen2021finqa, lu2023dynamic} and multi-step analysis \cite{zhang2023crt, wu2025tablebench}, surpassing the simpler open-domain queries of NQ-TABLES \cite{herzig2021open, kweon2023open}. However, a key limitation persists: a reliance on static, single-source data. This is evident even in benchmarks like RETQA \cite{wang2025retqa}, which introduced SLU but remained confined to a static database. The gap is particularly salient in geospatial QA. MapQA \cite{li2025mapqa}, a large-scale improvement over prior map datasets \cite{punjani2018template, kefalidis2023benchmarking}, still overlooks the dynamic nature of map problems where values like commute times are not fixed. Our dataset's positioning is detailed in Table \ref{tab:cmp}.

\subsection{Methods for Complex TQA}
Methodologically, the paradigm has shifted from bespoke trained models \cite{li2023toward} to LLM-driven agents augmented with tools \cite{qin2026large}. While early tool use included calculators \cite{zhu2024tat} or Python's Pandas library \cite{zhang2023crt}, these are limited in scope. More advanced agent frameworks like MACT \cite{zhou2025efficient} demonstrate complex collaboration, but their reliance on Pandas for data manipulation introduces significant memory consumption and scalability issues. Crucially, existing Map QA methods, whether retrieval-based or Text-to-SQL \cite{li2025mapqa}, are inherently tethered to the static data limitations of the datasets they target.

To address these dual gaps, we introduce Re-CoQA, a large-scale complex map QA dataset featuring multi-source and dynamic queries. We also implement HIRE-Agent, a hierarchical multi-agent framework that serves as a strong baseline. Unlike prior work constrained by static data or scalability issues, HIRE-Agent leverages dedicated database and live map service APIs. This design enables it to query massive structured tables and access the real-time geospatial information essential for solving dynamic, real-world problems.

\section{ReCoQA Dataset}

\subsection{Data Collection and Preprocessing}
The foundation of our benchmark is a dataset of residential communities and POIs spanning eight major Chinese metropolises: Beijing, Guangzhou, Shenzhen, Suzhou, Hangzhou, Wuhan, Nanjing, and Tianjin.

\textbf{Community Data.} We adopt the collection methodology from \cite{wang2025retqa}, extracting attributes relevant to homebuyer decisions, including greening rate, average transaction price, property type, and sales status. The textual address of each community is then geocoded into precise latitude and longitude coordinates using the Amap web service API\footnote{https://lbs.amap.com/api/webservice/guide/api/georegeo \label{lab:geo}}. This structured data is organized into city-specific tables (e.g., ``Table for Communities in Guangzhou'').

\textbf{POI Data.} For POI data, we manually collect six categories of public facilities relevant to homebuyers: schools, hospitals, supermarkets, shopping malls, parks, and public transit stations (subway and bus). Following geocoding via the same Amap API\footref{lab:geo}, we engage real estate professionals to annotate these POIs with a refined classification scheme (e.g., distinguishing primary school from secondary school within the education category). The resulting data, name, label, and coordinates are organized into city-specific POI tables (e.g., ``Table for POIs in Guangzhou'').

\textbf{Location Pair Data.} To facilitate efficient proximity queries, we pre-compute and store two types of location-pair data. First, we generate POI-Community pairs by pairing each POI with all communities located within a 3-kilometer radius. Second, we create Community-Community pairs by pairing each community with its neighbors within a 1-kilometer radius. This process yields lookup tables of proximate entities, indexed by city and captioned by type (e.g., ``Table for Communities around POIs in Guangzhou'').

The complete data, organized into these four table types per city, is imported into a PostgreSQL database. On average, across the eight cities, these tables contain 5,639 (Community), 4,007 (POI), 122,484 (POI-Community), and 179,441 (Community-Community) rows, respectively.

\subsection{Tool and Function Definition}
\label{sec:amap}
To address realistic homebuyer inquiries, access to the static community tables is insufficient, as many questions necessitate dynamic geospatial computations. To bridge this gap, we define a set of external tools that emulate the core capabilities of a commercial mapping service such as Amap\footnote{\url{https://lbs.amap.com/api/webservice/summary}}. These capabilities are exposed as callable functions, allowing an LLM to retrieve information not contained in the static database. We provide four principal functions:
\begin{itemize}
\item \textbf{Time Query:} Calculates the travel time between an origin and a destination via different transportation modes.
\item \textbf{Distance Query:} Calculates the travel distance between two points, including straight-line distance, walking distance, and driving distance.
\item \textbf{Surrounding POIs Query:} Retrieves a list of specified POIs within a certain radius of a given location.
\item \textbf{Rush Hour Query:} Provides the travel time for driving and public transportation during peak hours.
\end{itemize}
To increase task complexity and more rigorously evaluate planning and intent parsing, we deliberately consolidate several vendor-level API endpoints into a smaller set of more expressive, higher-level functions. This design compels the model to infer user intent and select appropriate parameters for each call. Together, these tools and the static tables constitute the complete information environment for our QA task.

\subsection{QA Pair Generation}
\label{sec:qa}
Building on the static tables and the defined tools, we construct a comprehensive set of QA pairs that reflect authentic homebuyer concerns. In consultation with experienced real-estate professionals, we identify 16 core information needs (e.g., amenity proximity, commute burden, and cross-property comparisons).

We instantiate these concerns as three progressively challenging question types:
\begin{enumerate}
\item \textbf{Simple Questions (Type 1):} Questions answerable via direct lookups in the static database.
\item \textbf{Compound Questions (Type 2):} Questions requiring lookups in database and calls to one or more of the predefined geospatial functions.
\item \textbf{Multi-step Reasoning Questions (Type 3):} Complex inquiries demanding a chain of database queries, function calls, and reasoning based on the function call results.
\end{enumerate}
We then author 41 query templates to systematically generate questions across these categories to get 45,040 initial queries. We provide more details about template filling and the example of our dataset in Appendix \ref{apdx:template}. To enhance linguistic diversity and better approximate natural language, we first use Qwen2.5‑72B \cite{Yang2024Qwen25TR} to paraphrase the template-based queries, followed by a verification step to ensure the integrity of key information slots. We then employ GPT‑OSS‑120B~\cite{agarwal2025gpt} to assess the naturalness of the queries before and after rewriting on a 0–5 scale, where lower scores correspond to more template-like expressions and higher scores correspond to more human-like phrasing. For human evaluation, we randomly sample 50 queries from both the original and rewritten sets and ask professional real estate salespeople to assign naturalness scores. Empirically, the average score assigned by GPT‑OSS‑120B increases from 3.99 to 4.24 after rewriting, while the average human score rises from 3.84 to 4.03, indicating a clear improvement in the naturalness and diversity of the resulting expressions. More detailed results are provided in Appendix~\ref{apdx:eva}.

\subsection{Tool Implementation via API Caching}
\label{sec:cache}
Executing the tool functions for the queries via live API calls would be prohibitively expensive and time-consuming. To ensure our benchmark is accessible, reproducible, and free for users, we adopt an API caching implementation strategy. We implement each function to query a local database instead of making a real-time network request. In this case, database acts as a cache, which we pre-populate by executing each unique API call required by our query corpus exactly once and storing its real-world result. This design allows for fast, cost-free, and deterministic execution of all tool calls. More details about API caching can be found in Appendix \ref{apdx:api_cache}.

\subsection{Data Annotation and Verification}
For each of the 45,040 queries, we construct a ground-truth execution trace comprising the ordered SQL statements and function calls necessary to derive the final answer. We then validate the executability of each trace, discarding any QA pair with a non-executable SQL statement or invalid function call. In addition, we apply a plausibility filter during instantiation to remove logically inconsistent queries (e.g., requesting walking time to a destination 20 km away). After filtering, we obtain 29,270 high-quality QA pairs.

Furthermore, we enrich these validated pairs with a comprehensive set of SLU labels. Then we manually map each query to one of 16 predefined intents and annotate 19 slot types using the standard IOB schema \cite{ramshaw1999text}. These annotations provide crucial supervisory signals for models learning to parse user requests. For each QA pair, we also provide comprehensive information including: the ground truth SQL statement and corresponding result, the tool API with parameters, the canonical answer, and a natural language answer. The rich annotation facilitates a wide spectrum of analyses and supports various downstream tasks.

Finally, to create the evaluation splits, we perform stratified sampling on the 29,270 pairs based on their original query templates. This strategy ensures a balanced distribution of question complexity across the training, validation, and test sets, which are partitioned in an 8:1:1 ratio. Please refer to Appendix \ref{apdx:statistics} for more statistical information.


\section{HIRE-Agent}
\subsection{Overview}
\begin{figure*}[ht]
\centering
\includegraphics[scale=0.45]{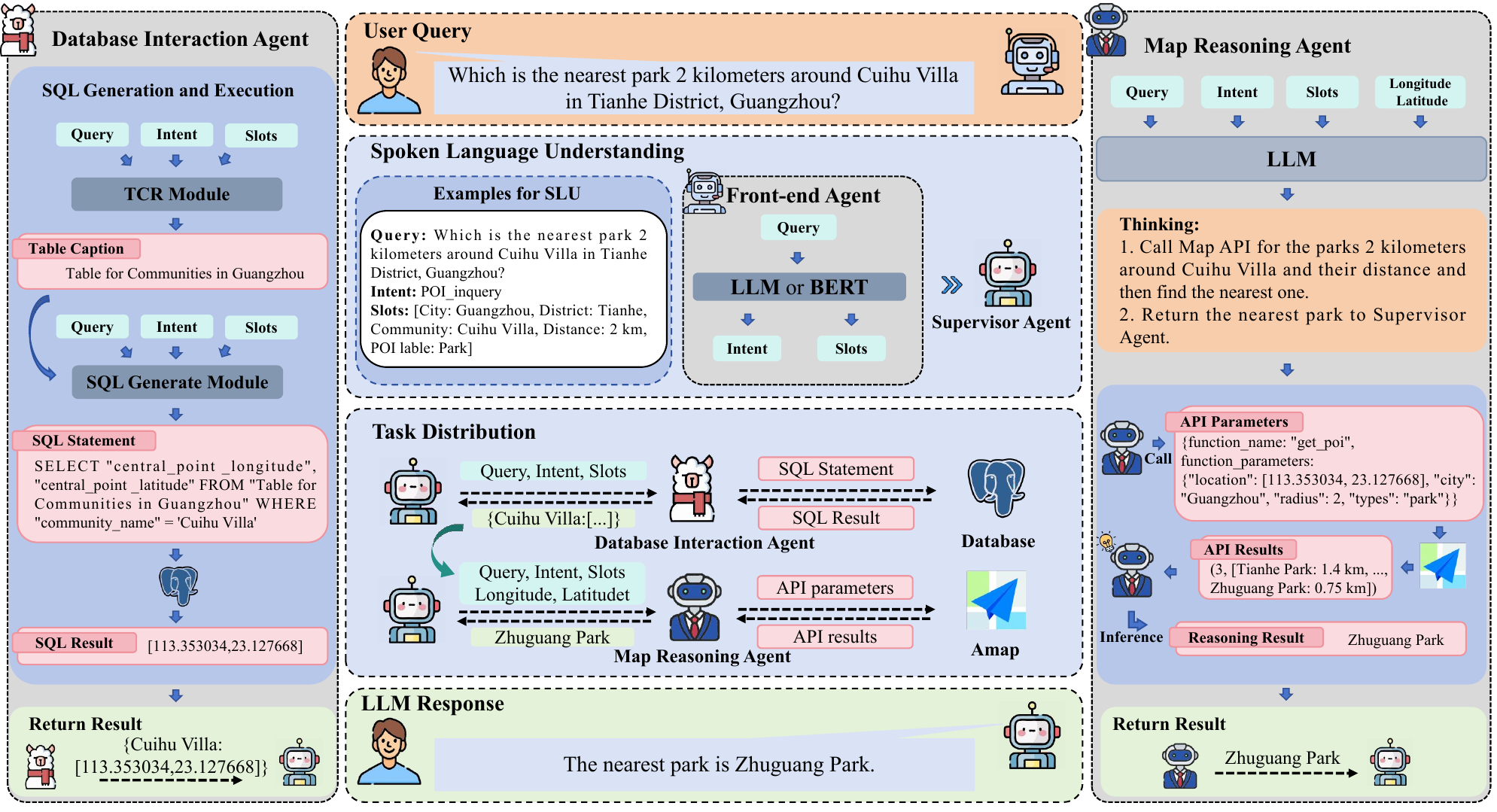}
\caption{The overall process of HIRE-Agent.}
\label{fig:overall}
\end{figure*}

As illustrated in Figure \ref{fig:overall}, our HIRE-Agent framework employs a hierarchical multi-agent architecture. The process begins with a Front-end Agent that extracts intents and slots from the user's query. This structured output is then passed to a central Supervisor Agent, which orchestrates the task fulfillment. The Supervisor decomposes the user's request into sub-tasks, delegates them to function-specific Specialist Agents, and synthesizes their returned results into a final, coherent response. All system prompts guiding agent behavior are detailed in Appendix \ref{apdx:sys_prompts}.



\subsection{Front-end Agent}
\label{sec:frontend}
The Front-end Agent is tasked with performing spoken language understanding. This process involves two key sub-tasks: intent detection and slot filling, which collectively transform an unstructured user query into a structured representation. Following previous work \cite{wang2025retqa}, we fine-tune a BERT \cite{devlin2019bert} model to implement the SLU module and perform intent detection and slot filling. The resulting structured representation is then passed to the Supervisor Agent to initiate the task execution phase.


\subsection{Supervisor Agent}
The Supervisor Agent serves as the central orchestrator within the HIRE-Agent architecture, as depicted in the center of Figure \ref{fig:overall}. Its primary responsibilities include: ingesting the structured intent and slots from the Front-end Agent, dynamically decomposing the main task into a sequence of sub-tasks, dispatching these sub-tasks to appropriate specialist agents, gathering and synthesizing intermediate evidence returned by them, and ultimately formulating the final response.

Upon receiving the initial structured input, the Supervisor Agent initiates a planning phase. It leverages Chain-of-Thought (CoT) \cite{chen2025towards} prompting to generate an explicit execution plan. This reasoning process is guided by a system prompt that outlines the specific capabilities and responsibilities of each specialist agent. Based on this plan, the agent identifies the initial specialist required and dispatches the first sub-task, passing along the query, intents, slots, and any relevant context.

The subsequent process is an iterative loop governed by the feedback from specialist agents. A specialist's response determines the next action. Upon receiving a successful result, the Supervisor Agent evaluates if the collected evidence is sufficient to fulfill the original user request. If the condition is met, it proceeds to generate the final answer. Otherwise, it updates its context with the newly acquired information (e.g., geographical coordinates) and dispatches the next sub-task in its plan to the relevant specialist agent. Conversely, in the case of an error report or a declaration of inability, a replanning mechanism is triggered. The Supervisor Agent then revisits its strategy, re-evaluates the execution plan in light of the failure, and may reassign the task, possibly to a different specialist or with modified parameters. To prevent infinite loops, this iterative process terminates if a solution cannot be found within a predefined number of execution steps, at which point the agent reports that the query cannot be answered.

\subsection{Specialist Agent}
Specialist agents are expert modules designed to execute concrete sub-tasks delegated by the Supervisor Agent. Each agent is equipped with a specific set of tools and is responsible for performing its designated function and returning the results as structured evidence. In this paper, we develop two distinct specialist agents: a Database Interaction Agent and a Map Reasoning Agent.
\subsubsection{Database Interaction Agent}
The Database Interaction Agent is engineered for complex database querying and is composed of two synergistic modules: a Table Caption Retrieval (TCR) module and a SQL Generation module. Both modules leverage the in-context learning (ICL) capabilities of LLMs, each prompted with five curated examples (detailed in Appendix \ref{apdx:usr_prompts}).

The agent's workflow, illustrated on the left of Figure \ref{fig:overall}, is initiated upon receiving a task. The primary challenge is to identify the correct table schema for the query. To solve this, the TCR module employs a sophisticated two-stage process. First, it utilizes an LLM to transform the user's query, intent, and slots into a hypothetical summary of an ideal table caption. This generated summary essentially serves as an optimized, high-quality query. Subsequently, this summary is passed to a BM25 retriever \cite{robertson2009probabilistic}, which then searches and ranks the actual table captions in the database, retrieving the one most semantically similar to the generated summary.

With the most relevant table schema(s) identified, the SQL Generation module takes over. It synthesizes the original query, intent, slots, and the retrieved table caption to generate a precise SQL statement. This statement is then executed against the database. If the execution is successful, the query results are returned to the Supervisor Agent as evidence. For results containing geographical coordinates, corresponding location names are also extracted and returned. If the SQL execution fails, an error report is sent instead.



\subsubsection{Map Reasoning Agent}
The Map Reasoning Agent is specialized in handling geospatial queries by interfacing with Amap API tools. As depicted on the right of Figure \ref{fig:overall}, it receives the user query, intents, slots, and any previously gathered geographical coordinates from the Supervisor Agent.

Its process begins with a tool selection phase. The agent first reasons about the most appropriate tool to use from its available toolkit (listed in Section \ref{sec:amap}) based on the task requirements and internal tool descriptions (see Appendix \ref{apdx:tool} for examples). After selecting a tool, it determines and generates the necessary parameters for the API call. Following the tool invocation, the agent evaluates the returned result. If the result directly answers the query, it is passed back to the Supervisor Agent. For queries that require additional synthesis (e.g., comparing the distance from two retrieved points), the agent performs a final reasoning step to infer the answer before returning it. The agent reports an error to the Supervisor under two specific conditions: first, if the geographical coordinates required for the task are missing, insufficient, or erroneous; and second, if it cannot derive a conclusive answer from the tool's output within a limited number of attempts.



\begin{table*}[htbp]
\centering
\scalebox{0.85}{
\begin{tabular}{cccccccccc}
\toprule
\multirow{2}{*}{Model} & \multirow{2}{*}{Method}  & \multicolumn{2}{c}{Type 1} & \multicolumn{2}{c}{Type 2} & \multicolumn{2}{c}{Type 3} & \multicolumn{2}{c}{Overall} \\ 
\cmidrule(r){3-4} \cmidrule(r){5-6} \cmidrule(r){7-8} \cmidrule(l){9-10}
&&F1 & Acc & F1 & Acc & F1 & Acc & F1 & Acc \\ 
\midrule
\multirow{2}{*}{Qwen2.5-72B} & Standard & 0.8082& 0.8082& \textbf{0.7229} & \textbf{0.7110} & 0.4011 & 0.3973 &0.5905 &0.5855  \\ 
 & HIRE-Agent & \textbf{0.8862} & \textbf{0.8862} & 0.6701&0.6581 & \textbf{0.6211}&\textbf{0.6211} & \textbf{0.7021}&\textbf{0.6989} \\
 \midrule
\multirow{2}{*}{Qwen3-8B} & Standard &0.6878 &0.6878 & 0.5176&0.5148 & 0.2649&0.2657 & 0.4405&0.4393 \\
 & HIRE-Agent & \textbf{0.8188}&\textbf{0.8188} & \textbf{0.7570}&\textbf{0.7484} & \textbf{0.5489}&\textbf{0.5489} & \textbf{0.6730}&\textbf{0.6707} \\ 
 \midrule
\multirow{2}{*}{Qwen3-30B A3B} & Standard & 0.7394&0.7394 & 0.5944&0.5871& 0.3531&0.3512& 0.5159& 0.5131 \\ 
 & HIRE-Agent & \textbf{0.7659}&\textbf{0.7659} & \textbf{\underline{0.8748}}&\textbf{\underline{0.8645}} & \textbf{\underline{0.8384}}&\textbf{\underline{0.8371}} & \textbf{\underline{0.8294}}&\textbf{\underline{0.8260}} \\
 \midrule
\multirow{2}{*}{GLM4.5-Air} & Standard & 0.8611&0.8611 & 0.6346&0.6232 &0.4112 &0.4093 & 0.5856&0.5817 \\ 
 & HIRE-Agent & \textbf{\underline{0.9101}}&\textbf{\underline{0.9101}} & \textbf{0.8013}&\textbf{0.7923} & \textbf{0.6075}&\textbf{0.6069} & \textbf{0.7363}&\textbf{0.7336} \\ 
  \midrule
  \multirow{2}{*}{Average} & Standard &0.7741& 0.7741& 0.6174& 0.6090&0.3576 & 0.3559&0.5331 &0.5299 \\ 
 & HIRE-Agent & \textbf{0.8453}&\textbf{0.8453} & \textbf{0.7768}&\textbf{0.7658} & \textbf{0.6540}&\textbf{0.6535} & \textbf{0.7352}&\textbf{0.7323} \\ 
 \bottomrule
\end{tabular}}
\caption{Overall performance of three kinds of queries. The underline represents the best performance in each column and the bold font indicates the best performance of each model.}

\label{tab:main}
\end{table*}
\section{Experiments}

\subsection{Implementation and Metrics}
To evaluate the effectiveness of the proposed HIRE-Agent framework on the ReCoQA benchmark, we instantiate the agent with four open-source LLMs, i.e., Qwen2.5-72B \cite{Yang2024Qwen25TR}, Qwen3-8B, Qwen3-30B A3B \cite{yang2025qwen3}, and GLM4.5-Air \cite{zeng2025glm}. The Frontend Agent for SLU is implemented by fine-tuning a BERT model\footnote{\url{https://huggingface.co/google-bert/bert-base-chinese}} for 10 epochs with cross-entropy loss on the training set.

As ReCoQA is a newly proposed benchmark without directly comparable baselines, we construct a simplified variant retaining only the Supervisor Agent (\textit{Standard}), where database querying, API invocation, and reasoning are handled exclusively by the Supervisor. The backend consists of a PostgreSQL database and four external APIs (Time, Distance, POI, and Rush-Hour Query), as described in Section~\ref{sec:amap}.

End-to-end performance is evaluated using Accuracy (Acc) and F1-score (F1). Acc is a strict exact-match metric that requires all predicted items to exactly match the ground truth, while F1 computes the harmonic mean of precision and recall at the item level, granting partial credit for multi-item answers. Additional implementation details are provided in Appendix~\ref{apdx:implements}. 
\subsection{Main Results}
\label{sec:main}

As shown in Table~\ref{tab:main}, we report overall results and performance on the three question types. The hierarchical HIRE-Agent architecture consistently outperforms the single-agent baseline (Standard), where the Supervisor alone handles all tasks. On average across models, hierarchy improves overall Accuracy by 0.2024 and F1 by 0.2021. Among the models, Qwen3-30B A3B achieves the best overall results (0.8260 Acc, 0.8294 F1), while Qwen3-8B performs the worst (0.6707 Acc, 0.6730 F1).

As expected, performance degrades with increasing question complexity, and this degradation is substantially milder under the hierarchical setting than under Standard. On average, from Type 1 to Type 2 and from Type 2 to Type 3, the Accuracy of Standard drops by 0.1651 and 0.2531, and F1 by 0.1567 and 0.2598, respectively. In contrast, the hierarchical architecture shows smaller drops of 0.0795 and 0.1228 in Accuracy and 0.0685 and 0.1123 in F1. For Type 3 questions in particular, the hierarchical HIRE-Agent surpasses Standard by 0.2976 in Accuracy and 0.2964 in F1 on average, indicating clear advantages on long-chain reasoning problems.

An interesting anomaly is Qwen3-30B A3B: under the hierarchical setting it excels on complex questions but performs unexpectedly poorly on simple ones. We find its strong reasoning ability cause it to “over-think” straightforward queries, leading to weaker planning. We will discuss this phenomenon in the ablation study and a qualitative case study is provided in Appendix~\ref{apdx:case}.

We also observe a capability trade-off, as no model dominates across all types. GLM4.5-Air is strongest on the simplest questions (0.9100 Acc), whereas Qwen3-30B A3B leads on the hardest ones (0.8371 Acc). Overall, even state-of-the-art LLMs fail to maintain uniformly high performance on ReCoQA, underscoring both the difficulty of the benchmark and the need for more robust agentic reasoning abilities.

\subsection{Ablation Study}
\label{sec:ablation}

We conduct an ablation study on the HIRE-Agent framework to identify the sources of performance variation across LLM backbones and locate critical bottlenecks.

We first evaluate the contribution of the Front-end Agent by comparing HIRE-Agent with and without SLU labels. As shown in Table~\ref{tab:frontend}, the \textit{Vanilla} setting (without SLU) yields consistently lower accuracy across all question types, with particularly poor performance on Type~3 questions (0.2921 on average). Introducing SLU labels substantially improves performance; for example, the average accuracy on Type~3 increases by 0.3614. The intent labels facilitate task allocation and tool selection, while slot labels provide accurate attributes and values for SQL generation and API invocation. These results confirm that decoupling intent understanding from execution is essential in this domain.

\begin{table}[ht]
\centering
    \scalebox{0.6}{
    \begin{tabular}{c c c c c c c c c}
\toprule
\multirow{2}{*}{Model}  & \multicolumn{2}{c}{Type 1} & \multicolumn{2}{c}{Type 2} & \multicolumn{2}{c}{Type 3}\\ 
\cmidrule(r){2-3} \cmidrule(r){4-5} \cmidrule(l){6-7}
& Vanilla & FT & Vanilla & FT & Vanilla & FT\\ 
\midrule
Qwen2.5-72B
 & 0.6098 & \textbf{0.8862} & 0.1355 &\textbf{0.6581} & 0.1289 &\textbf{0.6211}\\ 

Qwen3-8B & 0.1786 & \textbf{0.8188} &0.1161 & \textbf{0.7484}&0.0305 & \textbf{0.5489}\\ 

Qwen3-30B A3B & 0.5701 & \textbf{0.7659}  &0.5948 & \textbf{0.8645}& 0.5582 & \textbf{0.8371}\\ 

GLM4.5-Air & 0.6124 & \textbf{0.9101}& 0.5188& \textbf{0.7923}& 0.4508& \textbf{0.6069}\\ 
\bottomrule
\end{tabular}}
\caption{Impact of Front-end Agent on the final result.}
\label{tab:frontend}
\end{table}





Then, we assess the Database Interaction Agent using Executable Code Ratio (ECR) and pass@1 following \cite{he2024text2analysis} (Table~\ref{tab:sql}). For Type~2 and Type~3 queries, which only require retrieving location coordinates, models achieve high average pass@1 (0.9593) and ECR (0.9810). In contrast, performance on Type~1 (Text-to-SQL) queries is lower (0.8955 pass@1 on average) due to complex multi-table joins and constraints. Notably, final answer accuracy for Type~1 is consistently 3--4\% lower than SQL pass rates, indicating error accumulation in subsequent stages. Qwen3-30B A3B exhibits an especially large drop (nearly 10\%), suggesting weaker agentic planning and execution despite strong core reasoning.

\begin{table}[ht]
\centering
    \scalebox{0.8}{
    \begin{tabular}{c c c c c c c}
\toprule
\multirow{2}{*}{Model}  & \multicolumn{2}{c}{Type 1} & \multicolumn{2}{c}{Type 2\&3} \\ 
\cmidrule(r){2-3} \cmidrule(l){4-5}
& pass@1 & ECR & pass@1 & ECR \\ 
\midrule
Qwen2.5-72B
 & 0.9153 & 0.9709 & 0.9625 & 0.9817 \\ 

Qwen3-8B & 0.8571 & 0.9497 & 0.9351 & 0.9575\\ 

Qwen3-30B A3B & 0.8677 & 0.9523 & \textbf{0.9707} & 0.9918\\ 

GLM4.5-Air & \textbf{0.9418} & \textbf{0.9709} & 0.9689 & \textbf{0.9931}\\ 
\bottomrule
\end{tabular}}
\caption{Effectiveness of Database Interaction Agent.}
\label{tab:sql}
\end{table}

We also evaluate the Map Reasoning Agent via API label accuracy in Table~\ref{tab:api}. Accuracy on multi-step Type~3 questions is markedly lower than on Type~2, revealing limitations in handling longer reasoning chains. Moreover, the gap between tool-calling accuracy and final answer accuracy for both types indicates deficiencies in post-tool information synthesis.

\begin{table}[ht]
\centering
    \scalebox{1}{
    \begin{tabular}{c c c c}
\toprule
Model & Type 2 & Type 3 & Overall \\ 
\midrule
Qwen2.5-72B & 0.8206&0.6756&0.7270 \\ 

Qwen3-8B & 0.7652 & 0.6820 & 0.7115\\ 

Qwen3-30B A3B & 0.8852&0.8796&0.8816\\ 

GLM4.5-Air & \textbf{0.9239} &\textbf{0.8839}&\textbf{0.8980} \\ 
\bottomrule
\end{tabular}}
\caption{Effectiveness of Map Reasoning Agent.}
\label{tab:api}
\end{table}


Next, to further investigate the task planning capabilities of the Supervisor Agent, we evaluate its planning accuracy, with the results detailed in Table \ref{tab:plan}. Notably, Qwen3-30B A3B achieves the highest overall performance (0.9259), demonstrating a significant advantage over other models, particularly on Type 3 problems (0.9199). Conversely, we observe that Qwen3-30B A3B exhibits relative weaknesses when planning for Type 1 problems (0.9259), while GLM4.5-Air struggles with Type 3 problems (0.7741). These specific deficiencies corroborate the ``model weakness'' phenomenon identified in the main results. Ultimately, these findings underscore that the planning proficiency of the Supervisor Agent plays a critical role in determining final performance.

\begin{table}[ht]
\centering
    \scalebox{0.83}{
    \begin{tabular}{c c c c c}
\toprule
Model & Type 1& Type 2 & Type 3 & Overall \\ 
\midrule
Qwen2.5-72B &\textbf{0.9537}& 0.9355& 0.8442&0.8964\\ 

Qwen3-8B &0.8611& 0.7548& 0.7401& 0.7751\\ 

Qwen3-30B A3B &0.9259& \textbf{0.9368}&\textbf{0.9199}&\textbf{0.9259}\\ 

GLM4.5-Air &0.9392& 0.8787& 0.7741&0.8440 \\ 
\bottomrule
\end{tabular}}
\caption{Result of Task planning.}
\label{tab:plan}
\end{table}

\begin{figure}[ht]
\centering
\includegraphics[scale=0.13]{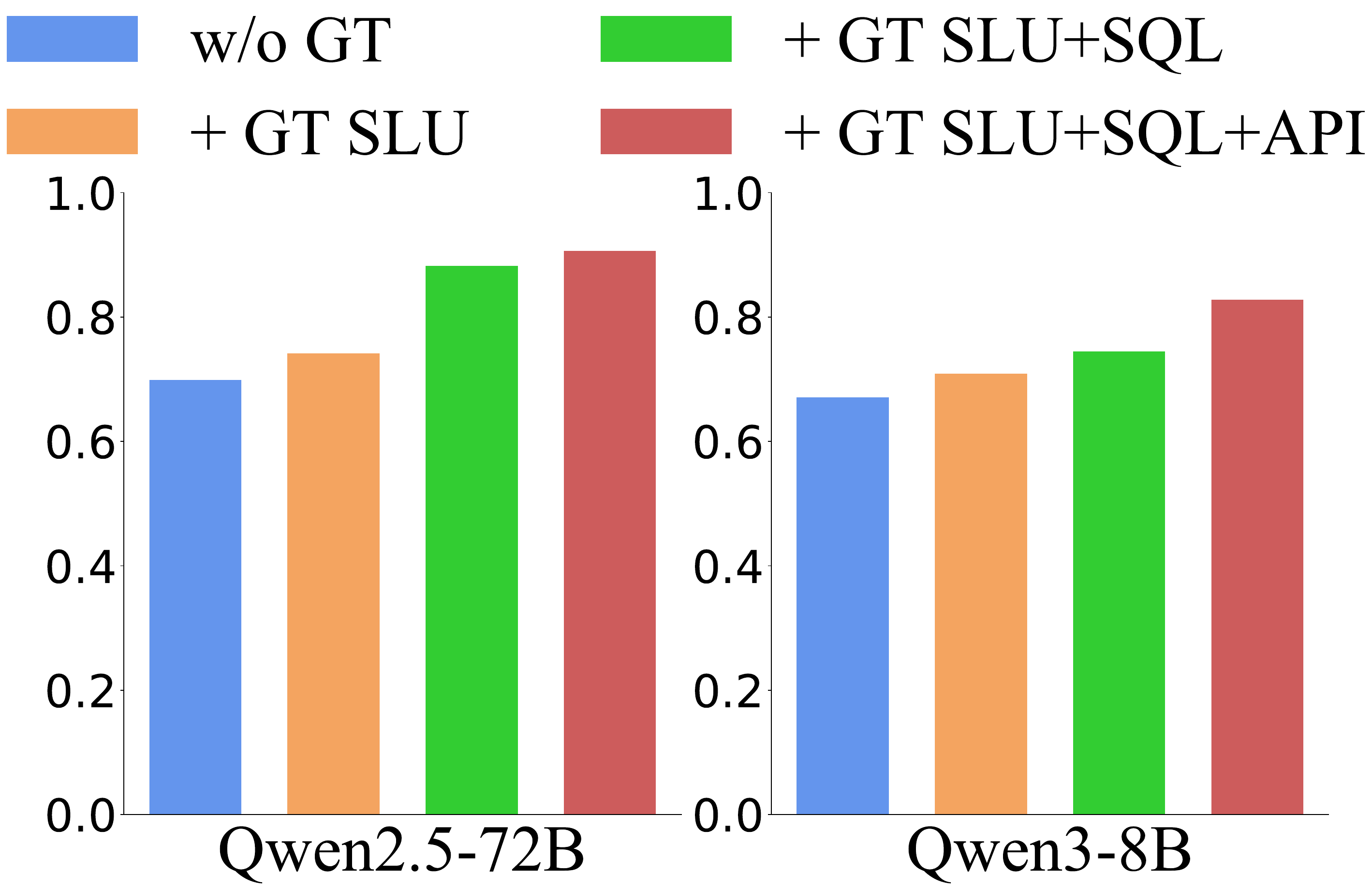}
\caption{The impact of different GT labels on the final result accuracy.}
\label{fig:gt_label}
\end{figure}

Finally, to holistically identify bottlenecks, we incrementally provide ground-truth (GT) labels for SLU, SQL, and API calls and measure the resulting accuracy gains (Figure~\ref{fig:gt_label}; Appendix~\ref{apdx:bottleneck}). All GT signals yield improvements, with GT SLU labels producing the smallest average gain (+0.0346), reflecting the strong performance of the BERT-based SLU model. Bottlenecks vary by backbone: for Qwen2.5-72B, GT SQL labels yield the largest gain (+0.1407), identifying SQL generation as the primary weakness, whereas for Qwen3-8B, GT API labels provide the largest boost (+0.0660), pointing to tool calling as the main limitation.

Crucially, even with all intermediate GT labels, average accuracy reaches only 0.8864. This reveals a final bottleneck, the \textit{synthesis gap} where errors persist in global planning and final reasoning despite perfect sub-agent execution. A representative case study is provided in Appendix~\ref{apdx:case}. This finding indicates that ReCoQA challenges not only individual agent competencies but also their ability to coordinate effectively.


\subsection{Real-world Questions Result}
To evaluate the performance in real-world scenarios, we collaborate with experienced real estate professionals to curate authentic homebuyer inquiries targeting information within our dataset (e.g., commuting duration, comparative pricing). We intentionally retain natural linguistic imperfections, such as syntax errors and colloquialisms, to reflect genuine user behavior without compromising clarity. Table~\ref{tab:real} summarizes query accuracy, highlighting Qwen3-30B A3B's top performance on Type 3 problems (0.5918) and GLM4.5-Air's on Type 1 (0.8800). All models show long-chain reasoning trends consistent with our template based questions. However, real-world phrasing and grammatical errors disrupt comprehension, resulting in comparable overall accuracy across all models (detailed in Appendix~\ref{apdx:real}).

\begin{table}[ht]
\centering
    \scalebox{0.83}{
    \begin{tabular}{c c c c c}
\toprule
Model & Type 1& Type 2 & Type 3 & Overall \\ 
\midrule
Qwen2.5-72B &0.8400& 0.8462& 0.5714&0.7100\\ 

Qwen3-8B &0.8400& 0.8462& 0.5714& 0.7100\\ 

Qwen3-30B A3B &0.8000& 0.8077&\textbf{0.5918}&0.7000\\ 

GLM4.5-Air &\textbf{0.8800}& 0.8077& 0.5714&0.7100 \\ 
\bottomrule
\end{tabular}}
\caption{Result of Real-world Questions.}
\label{tab:real}
\end{table}

\subsection{Evaluation of LLM-as-a-judge}
\label{sec:judge}
To broaden our evaluation, we employ the LLM-as-a-judge method \cite{gu2024survey} and apply GPT‑OSS‑120B as the judge to assess final responses (Table \ref{tab:judge}) of different LLMs. We analyze the judgment results and find that for questions with multiple correct answers (e.g., ``List all POIs...''), the order does not affect correctness, but the LLMs misjudge correct answers with reversed order. Furthermore, for ``how many'' questions, the correct answer can only be a specific number, while LLMs judge the answers of simple enumeration as correct. Given the precise data integration required by our dataset, we conclude that Exact Match remains the most reliable evaluation metric. Further case studies are detailed in Appendix \ref{apdx:case}.
\begin{table}[ht]
\centering
    \scalebox{0.7}{
    \begin{tabular}{c c c c}
\toprule
Model & Exact Match & LLM-as-a-judge & Difference \\ 
\midrule
Qwen2.5-72B & 0.6989&0.6992&+0.0003 \\ 

Qwen3-8B & 0.6707 & 0.6684 & -0.0023\\ 

Qwen3-30B A3B & 0.8260&0.8229&-0.0031\\ 

GLM4.5-Air & 0.7336 &0.7309&-0.0027 \\ 
\bottomrule
\end{tabular}}
\caption{The difference between LLM-as-a-judge and Exact Match, where the ``+'' and ``-'' represent ``increase'' and ``decrease''.}
\label{tab:judge}
\end{table}
\section{Conclusion}
In this paper, we introduce ReCoQA, a large-scale benchmark designed to foster research in tool-augmented, multi-step reasoning for real estate question answering. Grounded in realistic scenarios and featuring 29,270 question-answer pairs with verifiable intermediate steps, ReCoQA addresses the critical lack of specialized resources and establishes a challenging, standardized testbed for the community. Furthermore, we propose HIRE-Agent as a strong baseline. Our experiments not only validate the effectiveness of this approach but also highlight the benchmark's difficulty.

\section*{Acknowledgments}
This work is supported in part by the National Natural Science Foundation
of China (NSFC) under Grant 62272050 and the grant of Beijing Normal-Hong Kong Baptist University sponsored
by Guangdong Provincial Department of Education;  in part by Zhuhai Science-Tech Innovation Bureau under
Grant No. 2320004002772 and the Interdisciplinary Intelligence
Super Computer Center of Beijing Normal University (Zhuhai). The acquisition of real estate data is supported by Elmleaf Limited (Shanghai).

\section*{Limitations}
The map service employed in this study currently supports invocation only within a Chinese environment, and we therefore initially release ReCoQA in Chinese. Although an English version is planned for community use, we deliberately refrain from automated translation at this stage. Real estate entities (e.g., community names and POIs) often carry rich cultural attributes, such as historical references or allusions, where direct translation may lead to semantic distortion or loss of nuance. Such noise could propagate through the reasoning chain and compromise the rigor of QA evaluation. To mitigate this risk, we plan to engage professional translators to ensure a faithful English release in future work. While ReCoQA is instantiated in the real estate domain, its core challenges, hybrid reasoning over structured databases and dynamic external APIs reflects a domain-agnostic workflow common to many real-world applications, including travel planning, logistics, and healthcare.

To maximize instruction-following performance, our experiments employ Chinese system prompts aligned with the dataset’s native language. However, the underlying agent framework, particularly the logic for slot extraction and API routing, is designed to be language-independent. To facilitate understanding and future adaptation, we provide English translations of the system prompts and schema definitions in Appendix \ref{apdx:sys_prompts}.

The dataset is generated using templates, which is a common practice in existing large-scale QA benchmarks. To enhance linguistic diversity and reduce template artifacts, we apply LLM-based rewriting and evaluate question quality through both automatic scoring and human assessment. We further report and analyze results from 100 real-world scenarios collected from human queries, which provide additional evidence of the benchmark’s capacity to evaluate complex, practical reasoning.

Regarding the interaction format, the current dataset follows a one question, one answer setting. However, real-world real estate decision-making often involves iterative information gathering through multi-turn interactions. Extending the dataset to support multi-turn dialogue remains an important direction for future work.

Finally, we observe that small-parameter models generally perform poorly on ReCoQA, and some models exhibit limited planning capabilities. These observations highlight the difficulty of the benchmark and suggest opportunities for improving agent design in future research.


\bibliography{custom}

\appendix

\section{Template Filling}
\label{apdx:template}
\begin{figure}[ht]
\centering
\includegraphics[scale=0.28]{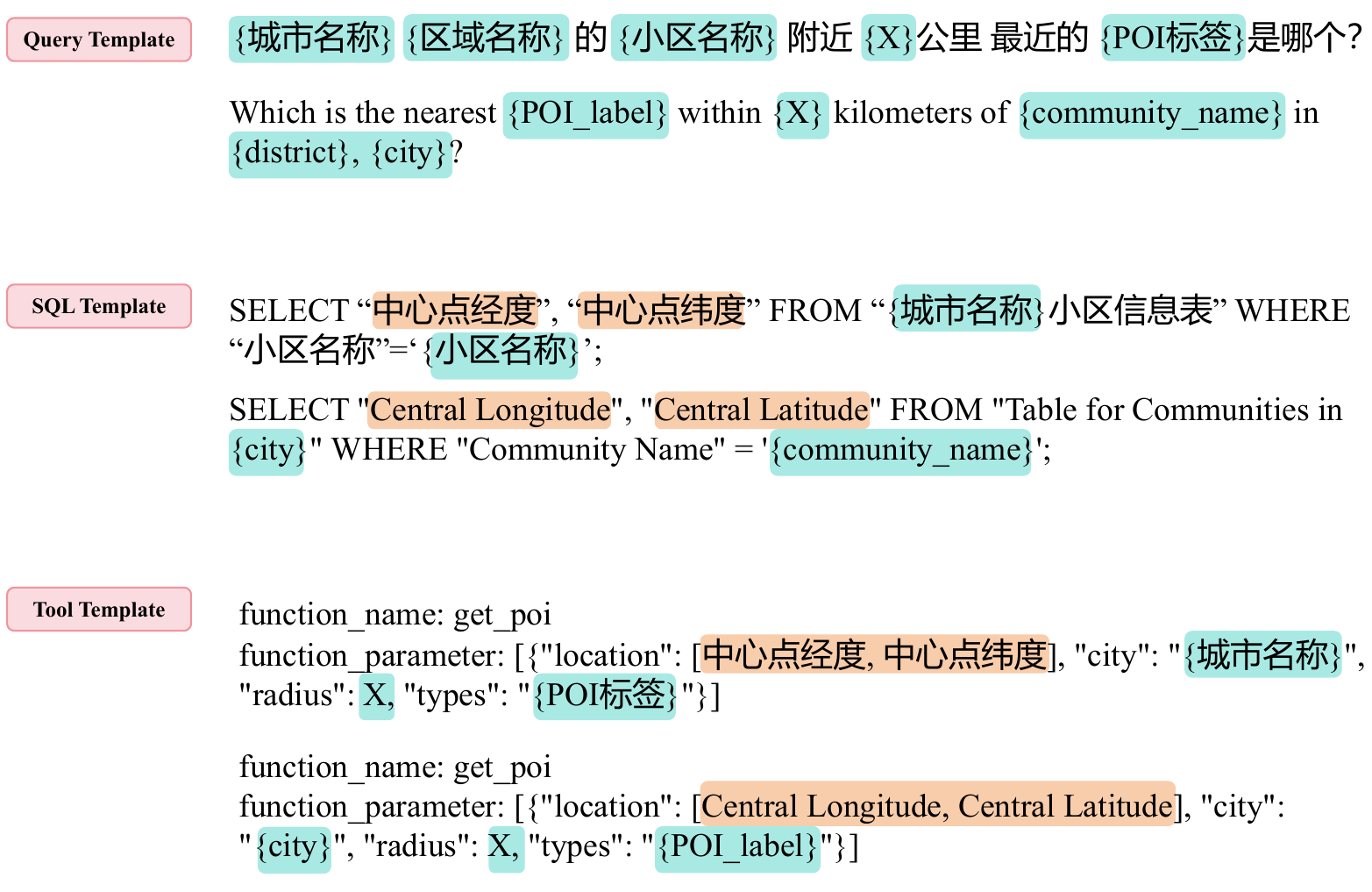}
\caption{Template for QA pairs generation.}
\label{fig:templates}
\end{figure}

We generate the QA pairs by filling templates described in section \ref{sec:qa}. As illustrated in Figure \ref{fig:templates}, each template pair consists of three essential components: a query template, a SQL template, and a tool template. These templates are parametrized using specific variables, which are sampled from the database. To generate a complete query and corresponding SQL statement, we randomly substitute values into the placeholder fields, such as \{\textit{community\_name}\}, \{\textit{POI\_label}\}, \{\textit{city}\}, and \{\textit{district}\}. Additionally, the parameter \{\textit{X}\} is stochastically generated, and its value is restricted to a maximum of 3 to regulate the amount of data returned after each API call. The SQL statement is subsequently executed to obtain the required longitude and latitude information (highlighted in yellow), which are then incorporated into the final API request parameters. Next, the interface is invoked with these parameters, and the resulting records, including the computed distances, are systematically archived. In cases where the returned results are empty, the corresponding queries are excluded from the dataset to maintain overall data quality.

\begin{figure*}[ht]
\centering
\includegraphics[scale=0.5]{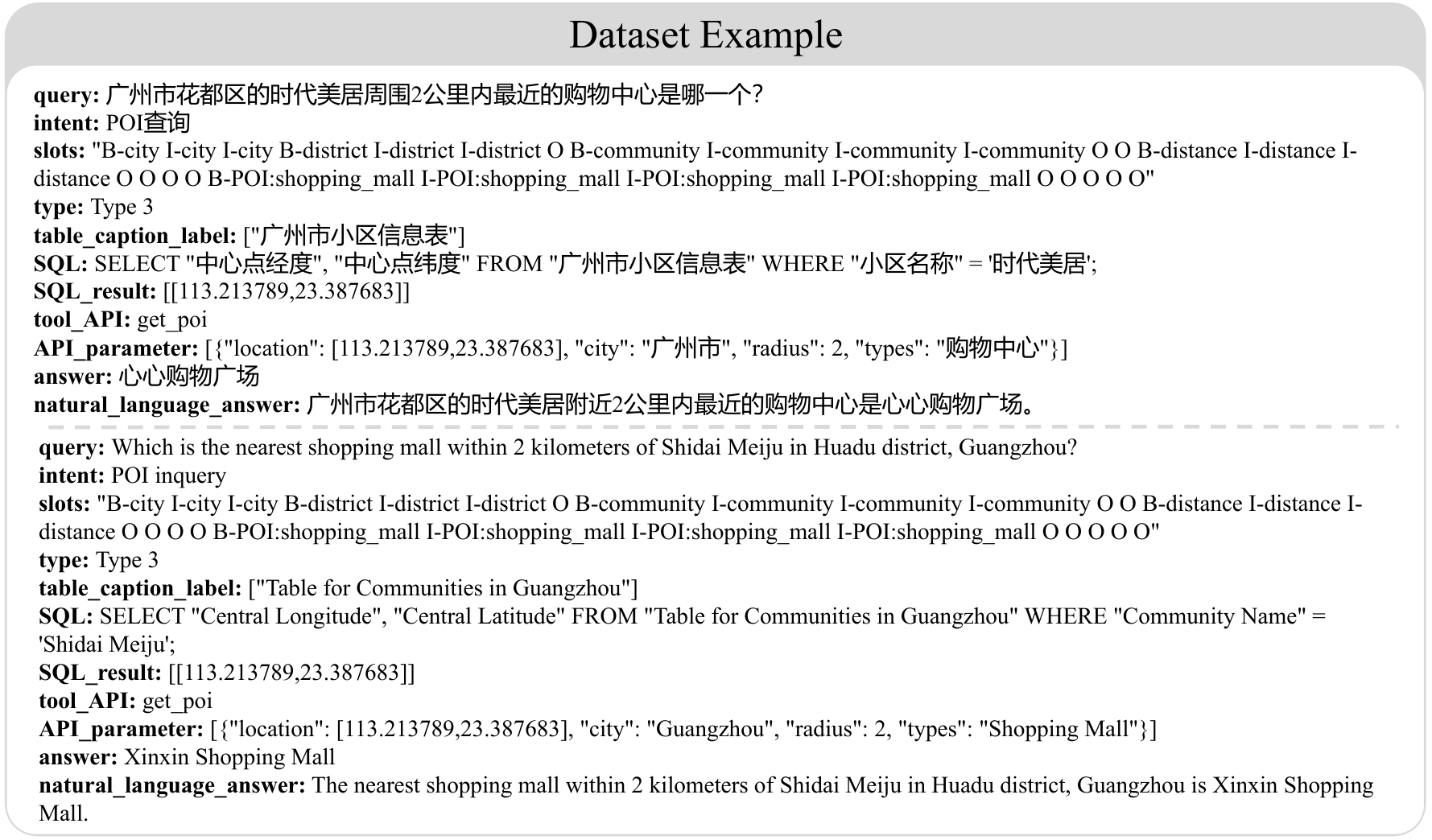}
\caption{One of the examples in the dataset.}
\label{fig:example}
\end{figure*}
After completing template instantiation, answer validation, and data annotation, we obtain the final QA pairs in the dataset. An example showcasing a filled QA pair constructed from the template is presented in Figure \ref{fig:example}. 


\section{QA Pairs Evaluation}
\label{apdx:eva}
\begin{figure*}[ht]
\centering
\includegraphics[scale=0.2]{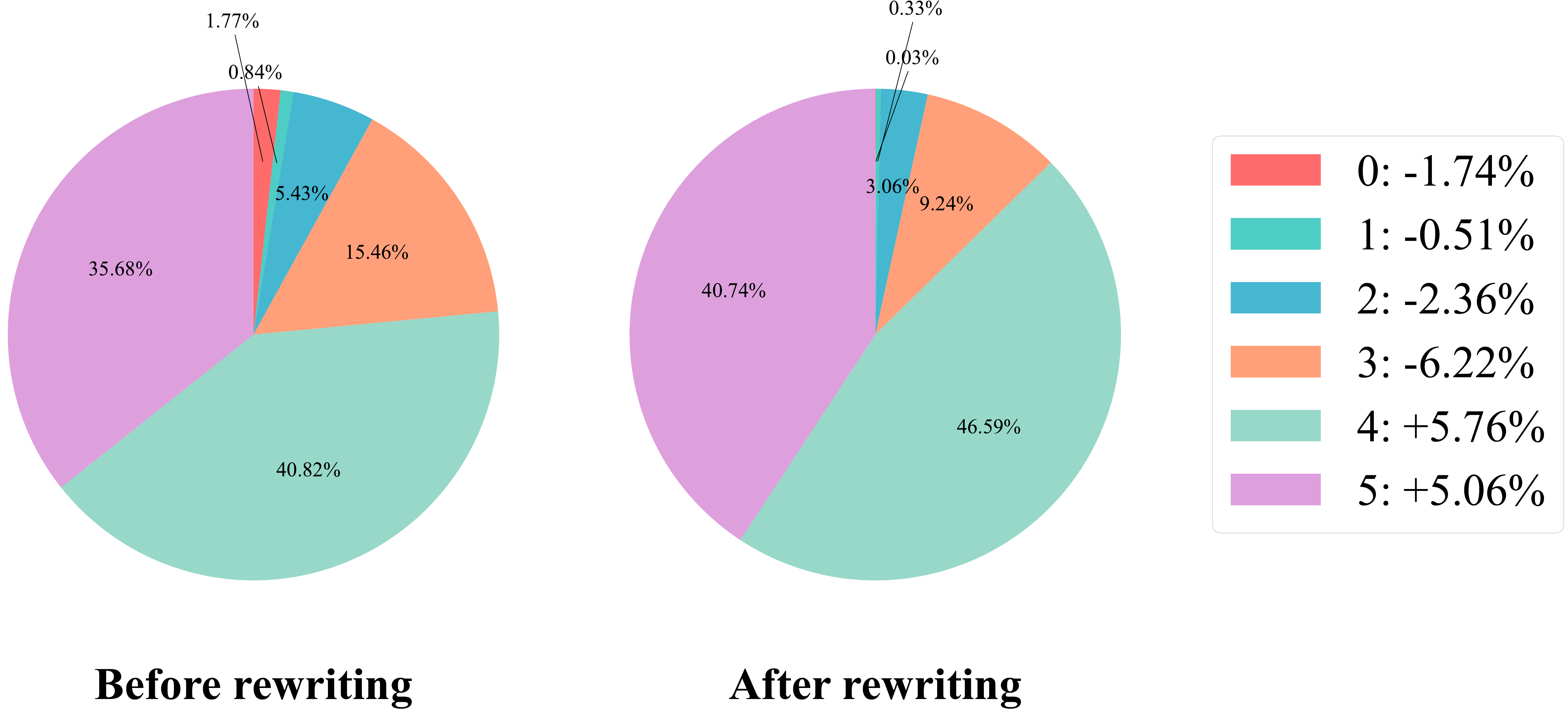}
\caption{The score of queries before and after rewriting, where the ``+'' and ``-'' represent ``increase'' and ``decrease''.}
\label{fig:rewrite}
\end{figure*}

We employ GPT-OSS-120B to quantify the naturalness of the QA pairs by grading their similarity to authentic human language. The statistical distributions of these scores, both prior to and following the rewriting process, are presented in Figure~\ref{fig:rewrite}. As indicated by the results, the prevalence of queries with a score of 0 denoting rigid, template-like structures decrease by 1.74\%. In contrast, the proportion of queries receiving scores of 4 and 5, which correspond to highly natural human expressions, increase by 5.76\% and 5.06\%, respectively. These shifts confirm that the rewriting process effectively enhances linguistic diversity and aligns the query phrasing more closely with human vernacular.

For the human evaluation component, we recruit salespeople from the real estate industry to assess a randomly sampled subset of questions from our dataset. Each sample comprise both the original and rewritten versions of a query. To ensure an unbiased assessment, the queries are shuffled and anonymized, establishing a blind evaluation setup where raters are unaware of the specific condition (original vs. rewritten) of each question. The raters are instructed to evaluate the naturalness of the queries using the same criteria applied in the LLM-based scoring. Upon calculating the mean scores for both conditions, we observe that the average human rating increase from 3.84 to 4.03, as noted in Section~\ref{sec:qa}. This significant improvement corroborates the efficacy of the rewriting mechanism in enhancing linguistic diversity and naturalness.


\section{API Caching}
\label{apdx:api_cache}
\subsection{Data Collection}
\label{apdx:collection}
As mentioned in Section~\ref{sec:cache}, we utilize the Amap API to acquire data concerning travel duration, distance, and surrounding POIs. Acknowledging that travel metrics are significantly influenced by traffic conditions and the chosen mode of transportation, we implement a time-stratified data collection strategy. Specifically, we collect commuting data for driving and public transportation during weekday peak hours (08:00 UTC+8) and off-peak hours (15:00 UTC+8). For scenarios that are less sensitive to real-time traffic fluctuations, we record baseline metrics for walking, driving, and cycling at midnight (00:00 UTC+8). However, given that public transportation is typically non-operational during late-night hours, we utilize data collected during the weekday off-peak window (15:00 UTC+8) as a representative proxy for these queries.

\subsection{SQL-style Interface}
\label{apdx:tool}
\begin{figure}[ht]
\centering
\includegraphics[scale=0.65]{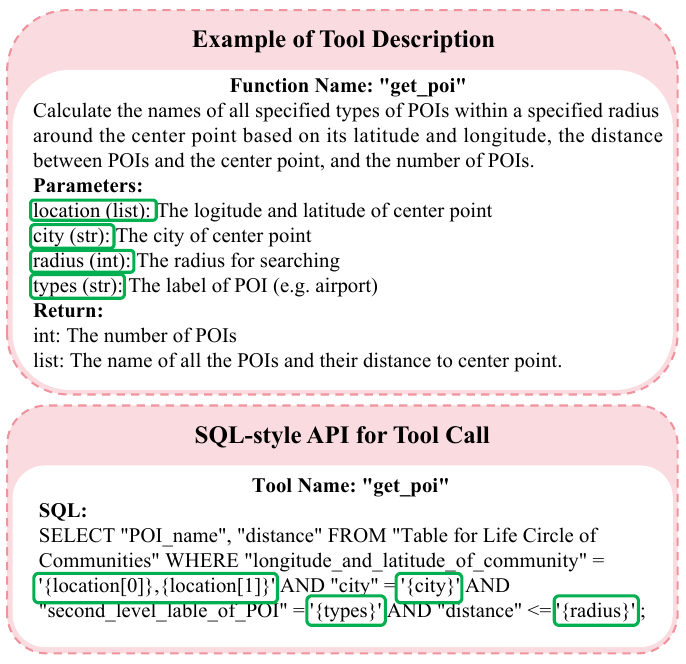}
\caption{Tool description and SQL-style API interface.}
\label{fig:tool}
\end{figure}

As described in Section \ref{sec:cache}, our approach leverages database querying to facilitate fast, cost-effective, and deterministic execution for all tool calls. Specifically, we first invoke the APIs to retrieve answers for every query and store these results in corresponding database tables. Subsequently, the relevant parameters acquired from these APIs are applied to our SQL-style interfaces, as illustrated at the bottom of Figure \ref{fig:tool}. This enables efficient offline caching and contributes to significant improvements in query response speed and resource utilization.

Finally, we provide the LLMs with details for each tool, including their descriptions, parameters with corresponding data types, and the output formats, as illustrated at the top of Figure \ref{fig:tool}. This comprehensive specification ensures that, as long as the LLMs invoke the appropriate parameters and adhere to the required data types, the SQL-style API interfaces will execute correctly and return the anticipated results.

\section{Statistics}
\label{apdx:statistics}
Table \ref{tab:statistics} summarizes the key statistical properties of our dataset. In this table, “Single Table” and “Multi Table” denote questions that require querying a single table or multiple tables, respectively. The “Slots” column indicates the types of slots present in the dataset, while “Avg. Slots” refers to the average number of slots per question. Notably, some questions may contain multiple instances of the same slot type. For example, the question “\textit{Which place takes the least amount of time to drive from Ziyue Chen Fu, Zhongjiao Yajun Chengdong Chunxiao, and Languang Yongjin Fu to Beichenglin Primary School in Tianjin?}” includes three different communities as input slots. For such cases, each occurrence is counted separately in the slot statistics.

\begin{table}[h]
\centering
\scalebox{1}{
\begin{tabular}{ll}
\toprule
Properties & Number \\
\toprule
Utterances & 29270 \\
Single Table & 6528\\
Multi Table & 22742\\
Single Intent & 24165\\
Multi Intent & 5105\\
Avg. Words per Utterance & 40.22\\
Intents & 16\\
Slots & 19\\
Avg. Slots & 4.86\\
Simple Questions & 7488\\
Compound Questions & 7766\\
Multi-step Reasoning Questions & 14016\\
\bottomrule
\end{tabular}}
\caption{Dataset Statistics.}
\label{tab:statistics}
\end{table}
\section{Database Interaction Agent Details}
\label{apdx:usr_prompts}

\begin{figure}[ht]
\centering
\includegraphics[scale=0.8]{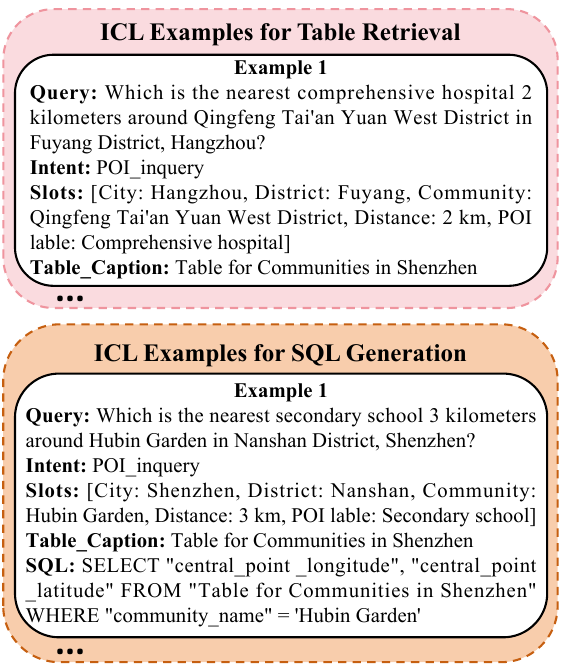}
\caption{ICL examples for Database Interaction Agent.}
\label{fig:ICL_SQL}
\end{figure}

For the Database Interaction Agent, we adopt the ICL paradigm for both table retrieval and SQL generation, as depicted in Figure \ref{fig:ICL_SQL}. In the table retrieval stage, each example comprises the user query, intent, slots, and the corresponding table caption. For SQL generation, we additionally include the ground-truth SQL statement for each instance. To enable effective learning, each prompt supplies the LLMs with five annotated examples to demonstrate the mappings for both table retrieval and SQL generation tasks.


\begin{table*}[htbp]
\centering
\scalebox{1}{
\begin{tabular}{ccccccccc}
\toprule
\multirow{2}{*}{Model} & \multicolumn{2}{c}{Type 1} & \multicolumn{2}{c}{Type 2} & \multicolumn{2}{c}{Type 3} & \multicolumn{2}{c}{Overall} \\ 
\cmidrule(r){2-3} \cmidrule(r){4-5} \cmidrule(r){6-7} \cmidrule(l){8-9}
&F1 & Acc & F1 & Acc & F1 & Acc & F1 & Acc \\ 
\midrule
Qwen2.5-72B & 0.8624&0.8624 & 0.6419 & 0.6297 & 0.5935 & 0.5935 & 0.6753 & 0.6721 \\ 
 \midrule
Qwen3-8B & 0.7645&0.7645 & 0.7131&0.7058 & 0.5467&0.5467 & 0.6465&0.6446 \\ 
 \midrule
Qwen3-30B A3B & 0.7778&0.7778 & 0.8811&0.8697&0.7967&0.7967 & 0.8141&0.8111 \\ 
 \midrule
GLM4.5-Air& 0.7791&0.7791 & 0.7625&0.7561 & 0.5673&0.5673 & 0.6731&0.6714 \\ 
  \midrule
Average & 0.7960&0.7960 &0.7497 & 0.7403& 0.6261& 0.6261&0.7023 &0.6998 \\ 
 \bottomrule
\end{tabular}}
\caption{performance of ICL method on three kinds of queries.}
\label{tab:icl}
\end{table*}

\section{Implements}
\label{apdx:implements}

The BERT model is implemented using PyTorch, while all agent modules are constructed utilizing the LangGraph and LangChain frameworks. The BERT model is trained and subsequently used for SLU labels prediction on a single GeForce RTX 4090 GPU. For HIRE-Agent framework, we set the max recursion number as 25 to avoid endless cycle of these agents.

To evaluate the performance of all LLMs, we employ a cluster of 19 NVIDIA A800-SXM4-80GB GPUs. Specifically, eight GPUs are allocated to the GPT- OSS- 120B, four to the Qwen2.5-72B model, four to the GLM4.5-Air model, one to Qwen3-8B, and the remaining two to the Qwen3-30B A3B model.

\section{ICL method Result}
\label{apdx:ICL}
We additionally employ an ICL strategy within the Front-end Agent for SLU labels prediction. Specifically, we construct the prompt by randomly sampling 26 examples from the training dataset, ensuring comprehensive coverage of the full spectrum of intents. These examples are then utilized to guide the LLMs in predicting SLU labels. During this experiment, all other components of the HIRE-Agent framework are held constant to isolate the impact of the ICL approach. The experimental results, presented in Table~\ref{tab:icl}, demonstrate that even with a limited number of annotated examples, the LLMs are capable of yielding competitive performance, highlighting the robustness of the ICL paradigm in low-resource scenarios.


\section{Ablation Study Details}
\subsection{Different Methods for Front-end Agent}
\label{apdx:frontend}
We evaluate the Precision (P), Recall (R), and F1 score of SLU labels prediction using BERT and ICL, as shown in Table \ref{tab:slu}.
\begin{table}[ht]
    \centering
    \scalebox{0.6}{
    \begin{tabular}{c c c c c c c}
\toprule
\multirow{2}{*}{Model} & \multicolumn{3}{c}{Intent} & \multicolumn{3}{c}{Slots} \\ 
\cmidrule(r){2-4} \cmidrule(l){5-7}
 & P & R & F1 & P & R & F1 \\ 
\midrule
Qwen2.5-72B & 0.9827 & 0.9827 & 0.9827 & 0.9475 & 0.9399 & 0.9437 \\ 
Qwen3-8B & 0.9412 & 0.9412 & 0.9412 & 0.9441 & 0.9200 & 0.9319 \\ 
Qwen3-30B A3B & 0.9820 & 0.9820 & 0.9820 & 0.9522 & 0.9460 & 0.9491 \\ 
GLM4.5-Air & 0.9806 & 0.9806 & 0.9806 & 0.9400 & 0.9260 & 0.9329 \\ 
BERT & \textbf{0.9997} & \textbf{0.9997} & \textbf{0.9997} & \textbf{0.9732} & \textbf{0.9576} & \textbf{0.9653} \\ 
\bottomrule
\end{tabular}}
\caption{Intent and Slots Prediction Results for ICL and BERT.}
\label{tab:slu}
\end{table}

The experimental results indicate that training a BERT model is the optimal approach for the Front-end Agent when a sufficiently large amount of data is available. Nevertheless, when only a limited number of SLU examples are provided, LLMs leveraging ICL can still accurately predict SLU labels. This finding demonstrates that ICL is a favorable alternative in scenarios where annotated data is scarce or manual labeling is impractical.

\subsection{Bottleneck Exploration}
\label{apdx:bottleneck}
As outlined in Section~\ref{sec:ablation}, we investigate the performance bottlenecks of the evaluated LLMs by assessing the accuracy of the final outputs under varying levels of GT guidance.

The detailed experimental results are summarized in Table~\ref{tab:bottleneck}.
\begin{table*}[htbp]
\centering
\begin{tabular}{cccccc}
\toprule
\textbf{Model} & \textbf{Query Type} & \textbf{w/o GT} & \textbf{GT SLU} & \textbf{GT SLU+SQL} & \textbf{GT SLU+SQL+API} \\
\midrule
\multirow{4}{*}{Qwen2.5-72B} & Type1 & 0.8624 & 0.9563 & 0.9656  & 0.9590 \\

 & Type2 & 0.6581 & 0.7652 & 0.8710 & 0.9819 \\

 & Type3 & 0.6211 & 0.6133 & 0.8435 & 0.8364\\
 & Overall & 0.6989 & 0.7414 & 0.8821 & 0.9062\\
\midrule
\multirow{4}{*}{Qwen3-8B} & Type1 & 0.8188 & 0.9021 & 0.9497 & 0.9378\\

 & Type2 & 0.7484 & 0.7845 & 0.8039 & 0.8916 \\

 & Type3 & 0.5489 & 0.5644 & 0.6020 & 0.7068 \\
  & Overall & 0.6707 &0.7091&0.7448& 0.8148\\
\midrule
\multirow{4}{*}{Qwen3-30B A3B} & Type1 & 0.7659 & 0.8849 & 0.9378 & 0.9365\\

 & Type2 & 0.8645 & 0.8735 & 0.8761 &0.9871 \\

 & Type3 & 0.8371 & 0.8534 & 0.8506 &0.8591 \\
  & Overall & 0.8260 &0.8668&0.8797&0.9127\\
\midrule
\multirow{4}{*}{GLM4.5-Air} & Type1 & 0.8453 & 0.9431 &0.9656  &0.9708 \\

 & Type2 & 0.7658 & 0.8077 & 0.9355 & 0.9858\\

 & Type3 & 0.6535 & 0.6154 & 0.8392 &0.8428 \\
  & Overall & 0.7336 &0.7503&0.8984 &0.9120\\
\bottomrule
\end{tabular}
\caption{Specific result for Bottleneck exploration.}
\label{tab:bottleneck}
\end{table*}
Generally, we observe a positive correlation between the granularity of provided GT labels and overall system accuracy. For Type 1 queries, performance reaches a plateau once GT SQL labels are supplied; while the subsequent inclusion of API labels introduces minor fluctuations, these remain within statistical error margins. In contrast, for Type 2 queries, the provision of GT API labels results in substantial performance gains, indicating that accurate tool invocation is the primary hurdle for this category. However, for Type 3 queries, the improvement remains marginal even with full GT supervision. This suggests that the bottleneck for these complex queries lies not in intermediate execution steps, but rather in the inherent reasoning capabilities of the models; even when correct API calls are guaranteed, logical errors during the final synthesis phase often lead to incorrect predictions.




\section{Case Study}
\label{apdx:case}
In this section, we elucidate specific observations and phenomena derived from the experimental results, adopting a Q\&A format to provide detailed explanations.

\textbf{Q1:} \textit{What accounts for the relatively weak performance of the Qwen3-30B A3B model on Type 1 queries?}

\begin{figure*}[htbp]
\centering
\includegraphics[scale=0.75]{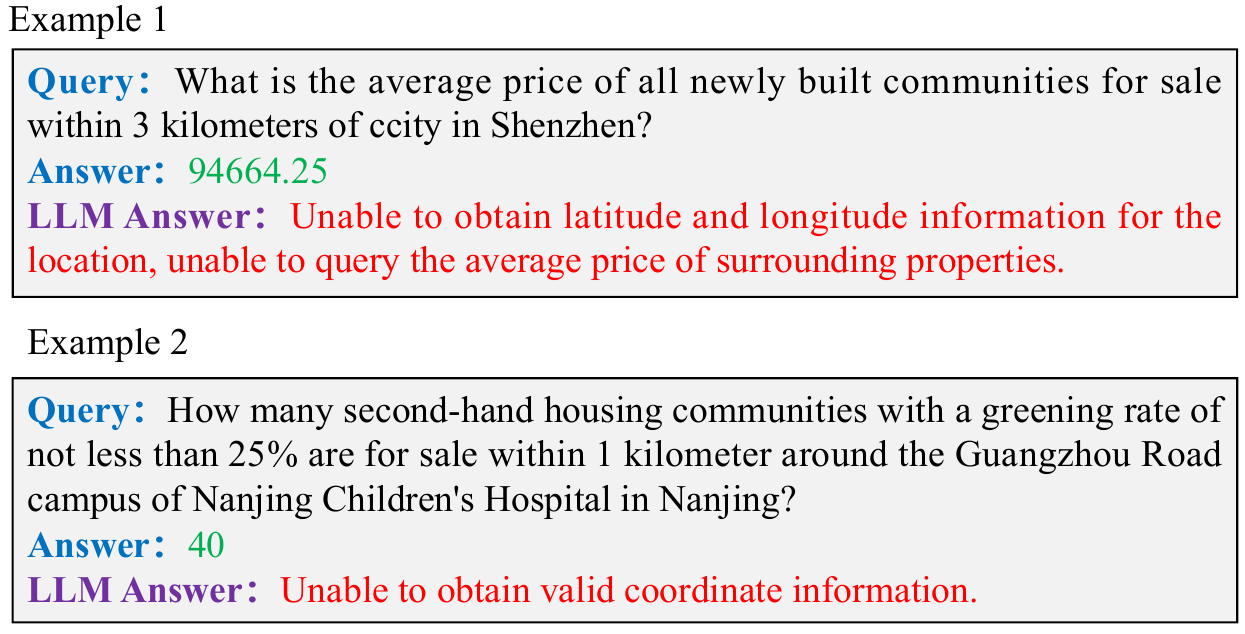}
\caption{Case Study for Qwen3-30B A3B.}
\label{fig:qwen3-30B}
\end{figure*}

\textbf{A1:} In the previous section, we analyze that the suboptimal performance of the Qwen3-30B A3B model on Type 1 questions primarily stems from its limited planning capability. To further validate this observation, we examine several representative cases exhibiting hallucinated answers. As illustrated in Figure~\ref{fig:qwen3-30B}, these cases correspond to Type 1 questions in our dataset, for which the correct answers can be obtained directly through database queries without requiring complex reasoning. For such questions, the appropriate planning and execution trajectory should follow the path ``Supervisor Agent $\rightarrow$ Database Interaction Agent $\rightarrow$ Supervisor Agent $\rightarrow$ User''. However, the Supervisor Agent erroneously incorporates the Map Reasoning Agent into the trajectory, even though the tools available to the Map Reasoning Agent are incapable of handling this class of queries and therefore return an inability signal to the supervisor. This misrouting ultimately causes the Supervisor Agent to hallucinate and fabricate answers instead of executing the correct database-only plan, thereby confirming that inadequate planning is a key bottleneck for the Qwen3-30B A3B model on Type 1 questions.

\textbf{Q2:} \textit{Why do LLMs fail to achieve optimal performance even when provided with full GT supervision?}

\begin{figure*}[htbp]
\centering
\includegraphics[scale=0.75]{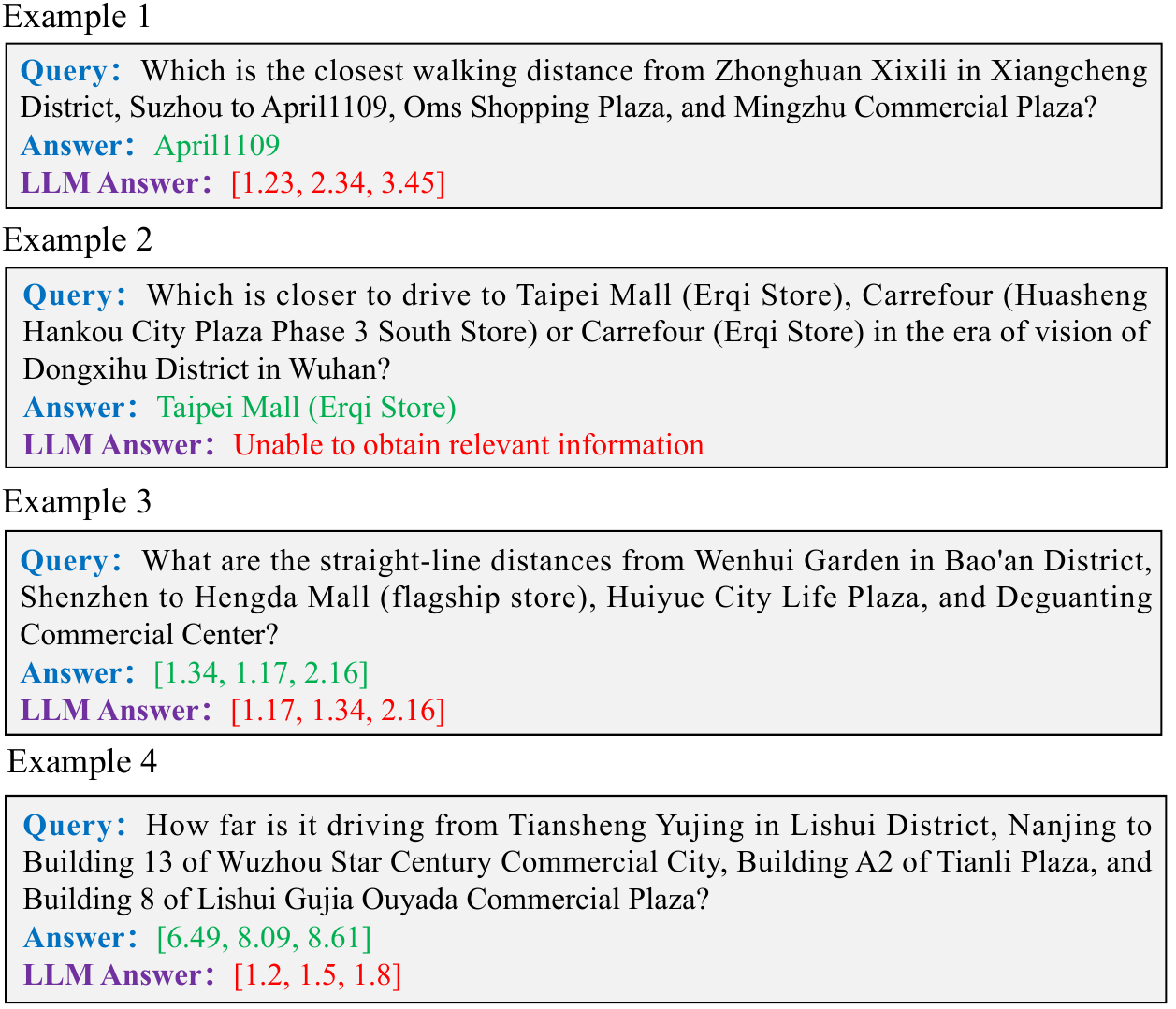}
\caption{Case Study for Qwen3-8B.}
\label{fig:qwen3-8B}
\end{figure*}

\textbf{A2:} We attribute this phenomenon to the models' insufficient planning and reasoning capabilities. To further substantiate this hypothesis, we conducted an in-depth analysis of the Qwen3-8B model, which exhibit the lowest performance in our experiments, as illustrated in Figure~\ref{fig:qwen3-8B}. 

Our investigation reveals that errors predominantly manifest in three aspects: (1) Hallucination induced by erroneous planning: As demonstrated in Examples 1 and 2, the Supervisor Agent fails to adhere to the optimal task allocation strategy due to flawed planning. Consequently, after receiving repeated negative feedback or failures from incorrectly invoked agents, the Supervisor Agent resorts to fabricating answers. For instance, in Example 2, the Supervisor Agent fails to incorporate the Database Interaction Agent into the execution plan, instead directly assigning the task to the Map Reasoning Agent. However, since the Map Reasoning Agent is unable to retrieve the requisite longitude and latitude coordinates without prior database lookup, it repeatedly returns error reports to the Supervisor. This persistent failure leads the Supervisor Agent to erroneously conclude that the necessary information is missing. (2) Information integration errors despite correct tool execution: As shown in Example 3, even when subordinate agents return accurate intermediate results, the Supervisor Agent introduces errors during the final response generation. Specifically, in this instance, the agent reverse the order of the retrieved information while integrating the final answer. (3) Complete hallucination: As depicted in Example 4, the Supervisor Agent occasionally generates responses that are entirely unrelated to the user query.


\textbf{Q3:} \textit{How to explain the difference between LLM-as-a-judge method and Exact Match?}

\begin{figure*}[htbp]
\centering
\includegraphics[scale=0.7]{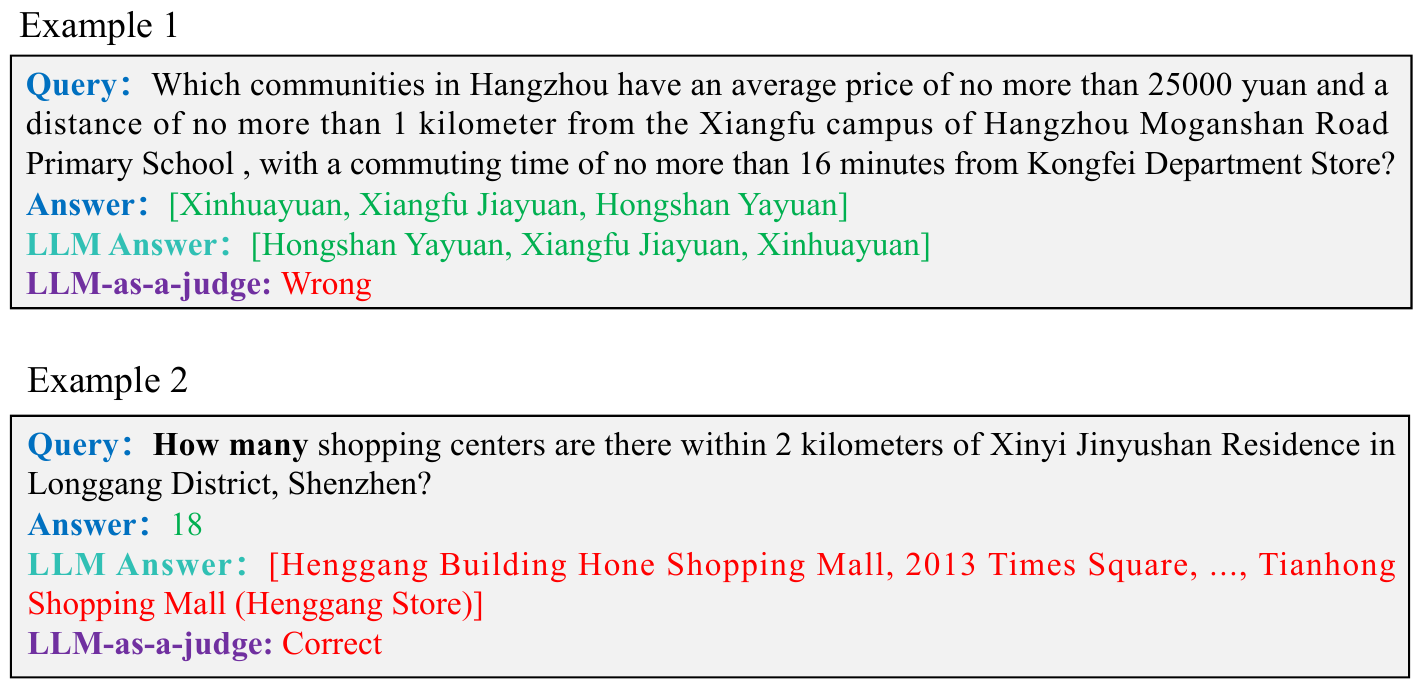}
\caption{Case Study for LLM-as-a-judge method.}
\label{fig:llm-judge}
\end{figure*}

\textbf{A3:} As discussed in Section \ref{sec:judge}, the discrepancies between the LLM-as-a-judge method and Exact Match primarily emerge in two types of queries: enumeration questions and ``how many'' questions. Figure \ref{fig:llm-judge} clearly illustrates these distinctions. For enumeration problem (Example 1), a response is valid as long as it includes all correct items, regardless of their order; however, the LLM-as-a-judge method incorrectly marks the valid response as wrong. Conversely, ``how many'' questions (Example 2) require a specific numerical answer, yet the LLM-as-a-judge method mistakenly accepts the answer that merely enumerates items without providing the explicit number. These consistent misjudgments across both problem types lead us to conclude that Exact Match is the best evaluation method.

\section{Real-world Questions Analysis}
\label{apdx:real}
\begin{figure*}[htbp]
\centering
\includegraphics[scale=0.7]{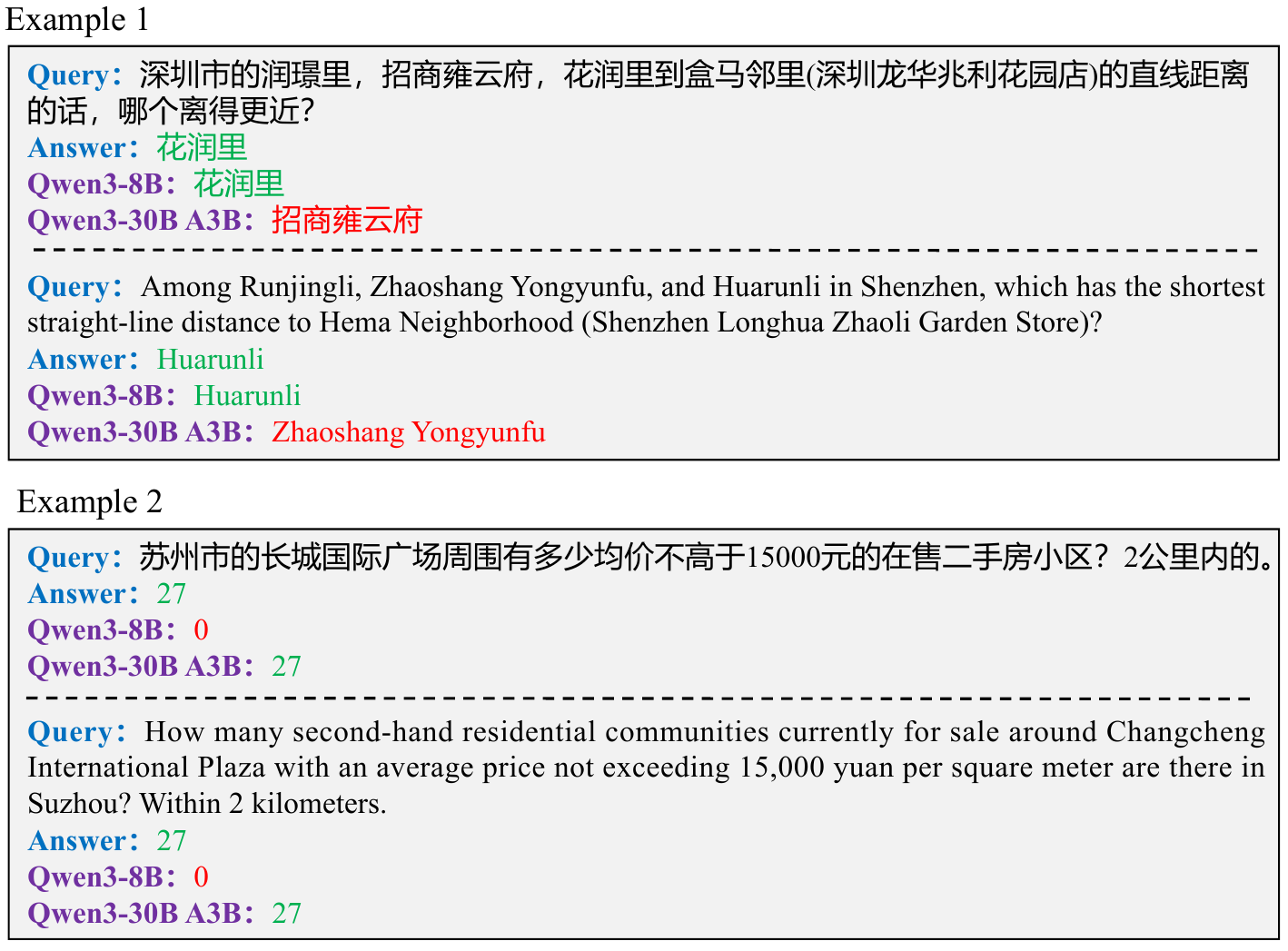}
\caption{The impact of real-world questions on different LLMs.}
\label{fig:real-world1}
\end{figure*}

To further evaluate how different LLMs handle real-world queries, we compare Qwen3-8B and Qwen3-30B A3B using some representative questions, as shown in Figure \ref{fig:real-world1}. Example 1 illustrates a scenario where users realize their query lacks context and append clarifications within the same sentence. While Qwen3-8B struggles with this irregular phrasing and yields incorrect answers, Qwen3-30B A3B interprets it successfully. Interestingly, this performance flips for certain reasoning tasks, where Qwen3-8B answers correctly but Qwen3-30B A3B fails. These findings demonstrate that real-world queries, particularly those containing syntax errors that do not hinder overall comprehension affect different LLMs to varying degrees.

\begin{figure*}[htbp]
\centering
\includegraphics[scale=0.7]{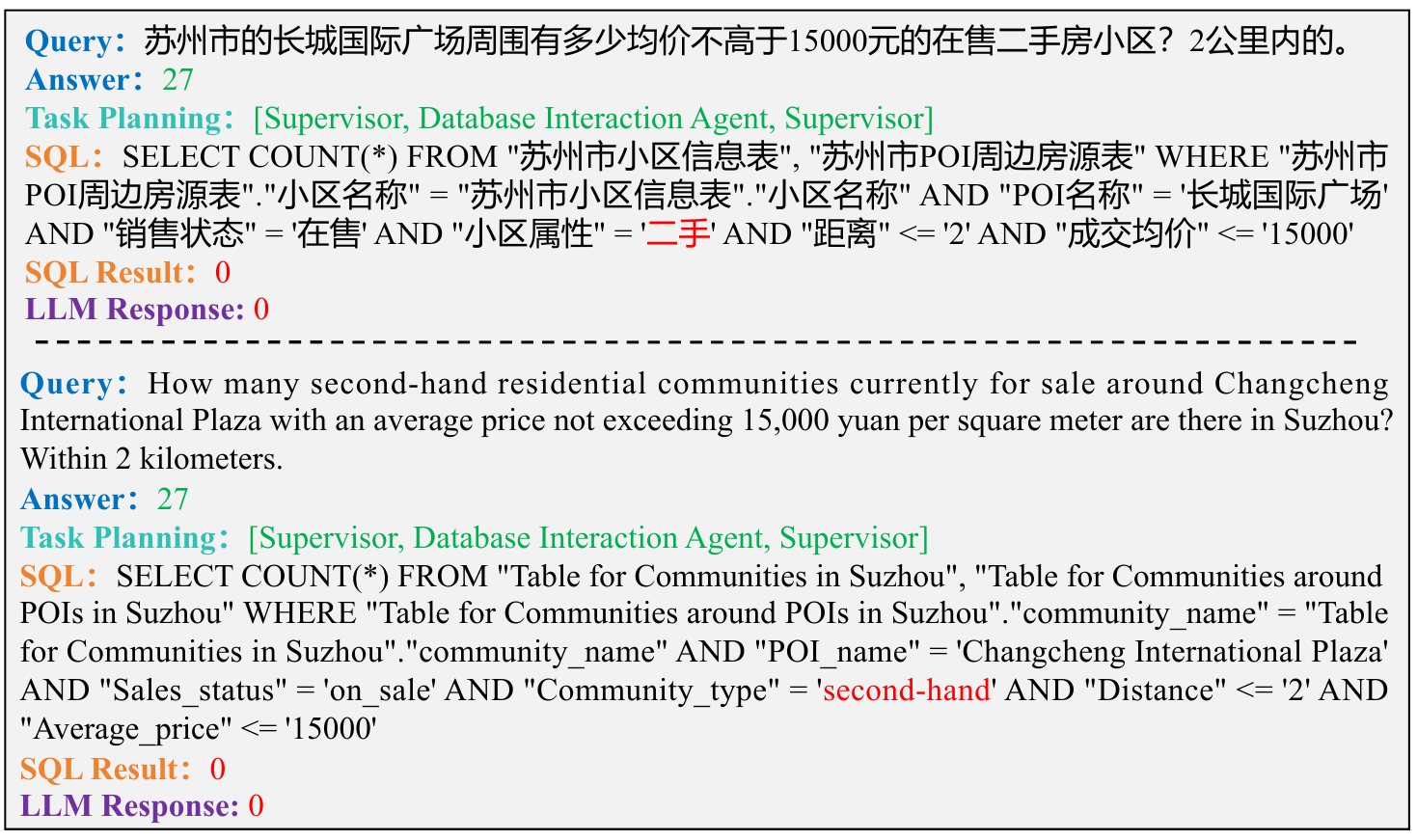}
\caption{The impact of real-world questions on planning and intermediate steps.}
\label{fig:real-world2}
\end{figure*}

To investigate how real-world queries affect task planning and its intermediate steps, we visualize the complete reasoning process of Qwen3-8B for Example 1, as shown in Figure \ref{fig:real-world2}. Although the model initially generates a correct task plan, it produces an inaccurate value during the database query phase. Specifically, it uses the term ``\textbf{second-hand}'' (marked in red) instead of the correct ``\textbf{second-hand property}''. Because this flawed SQL statement can still execute successfully, it retrieves erroneous data. The Supervisor Agent then mistakenly accepts this invalid result as correct and directly outputs the wrong answer.

\section{System Prompt}
\label{apdx:sys_prompts}
\begin{figure*}[ht]
\centering
\includegraphics[scale=0.5]{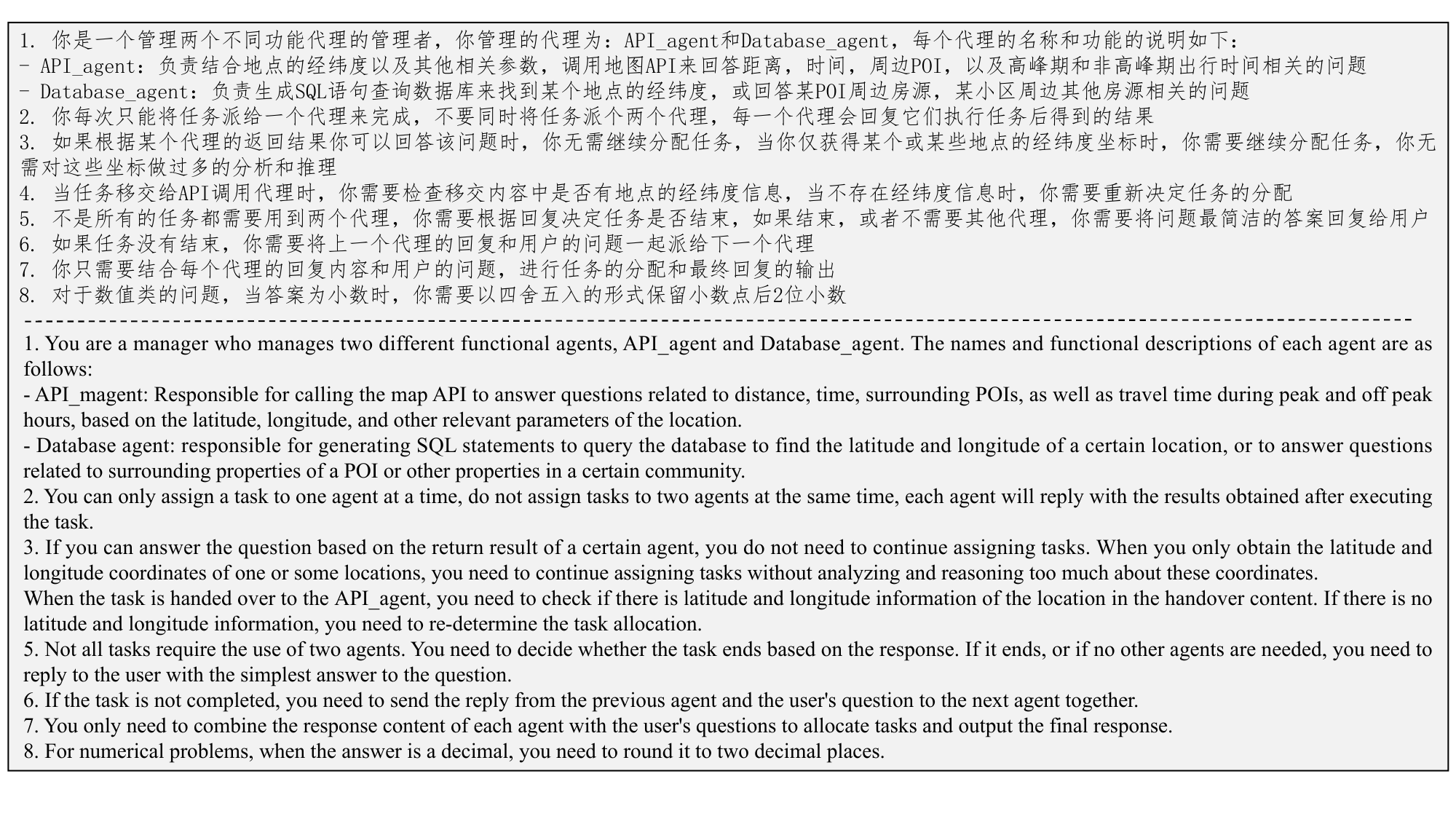}
\caption{System Prompt for Supervisor Agent.}
\label{fig:supervisor_prompt}
\end{figure*}

\begin{figure*}[ht]
\centering
\includegraphics[scale=0.55]{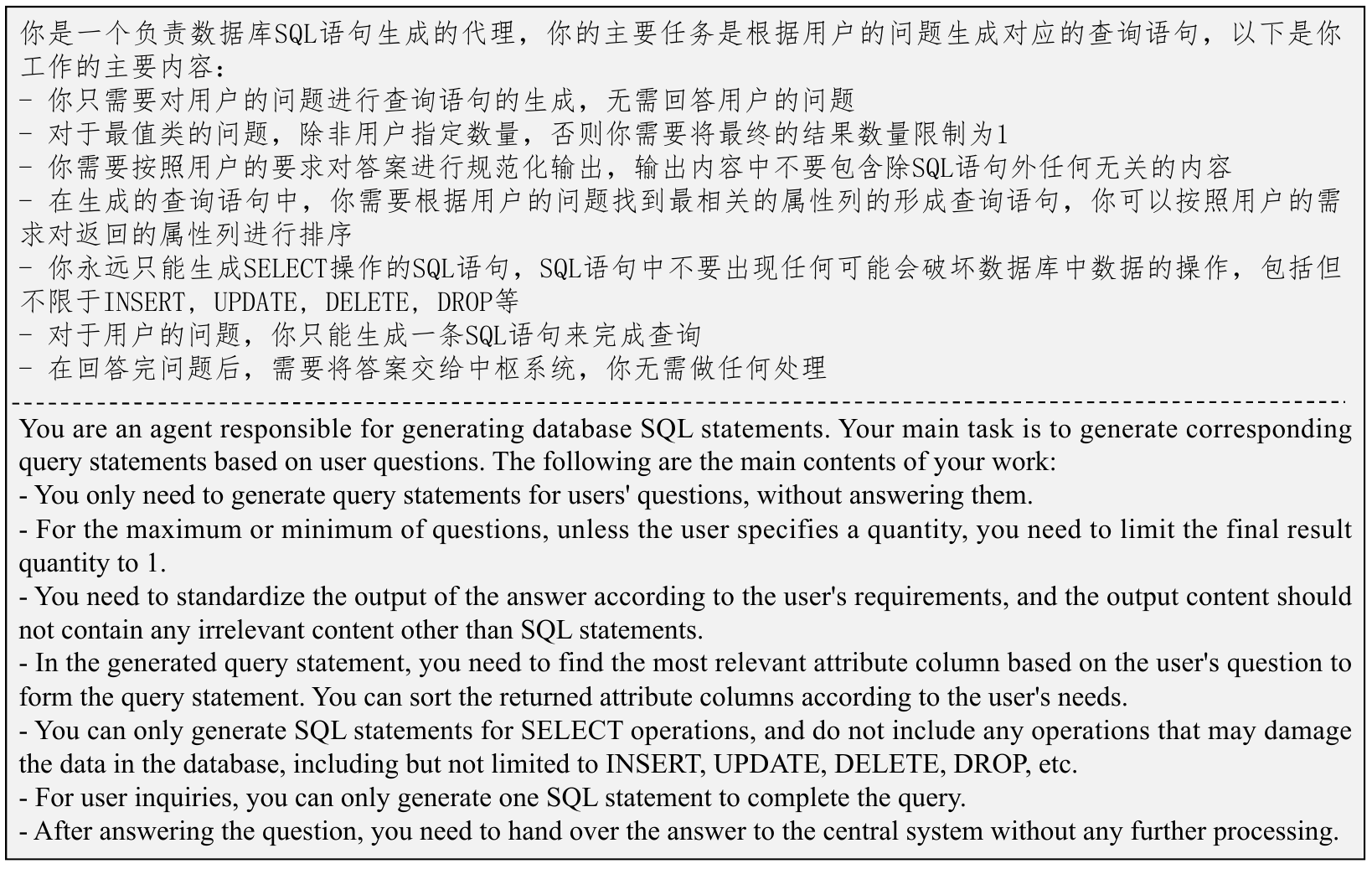}
\caption{System Prompt for Database Interaction Agent.}
\label{fig:database_prompt}
\end{figure*}

\begin{figure*}[ht]
\centering
\includegraphics[scale=0.55]{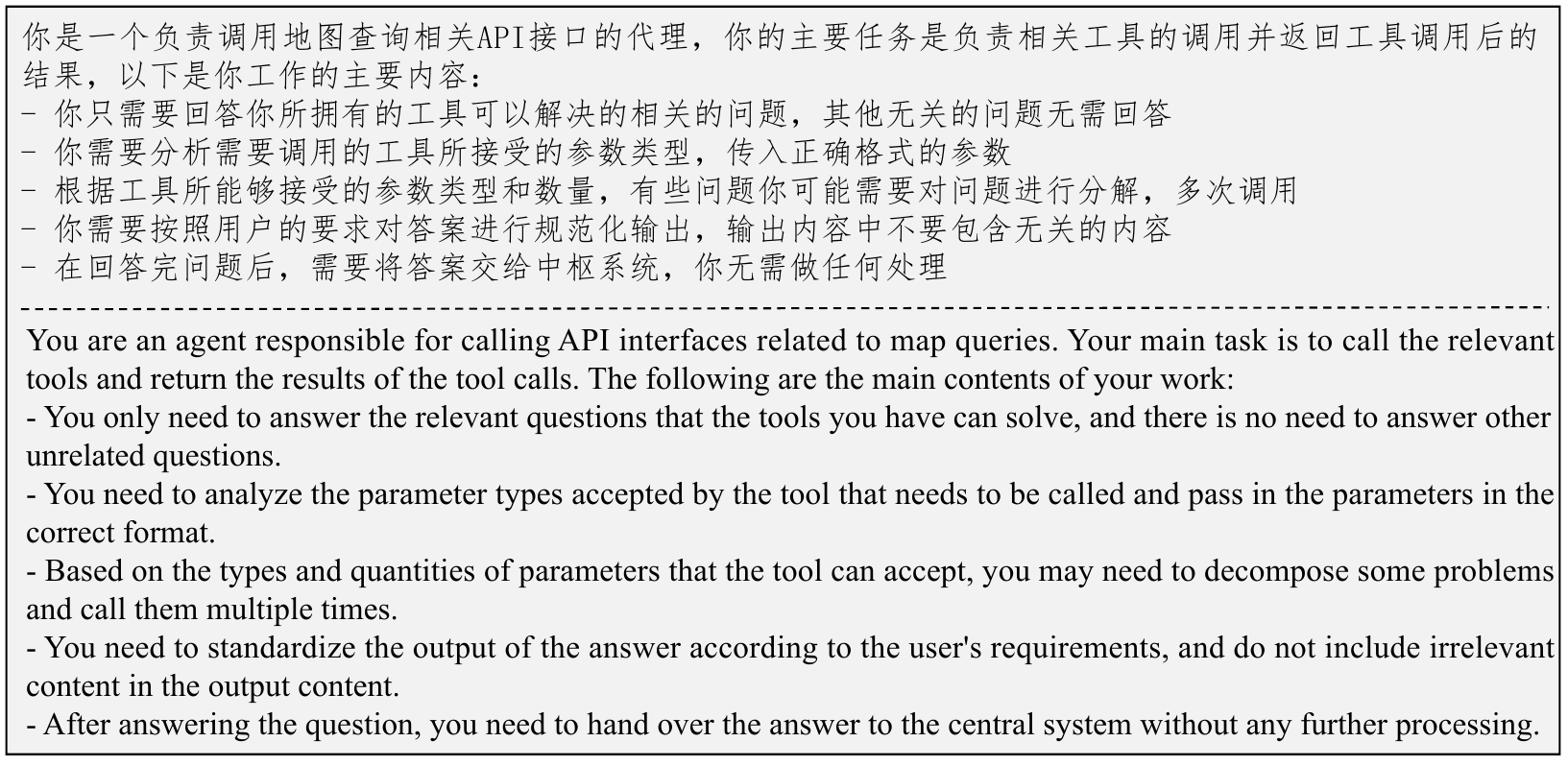}
\caption{System Prompt for Map Reasoning Agent.}
\label{fig:api_prompt}
\end{figure*}

The system prompts employed for the Supervisor Agent, Database Interaction Agent, and Map Reasoning Agent are illustrated in Figure \ref{fig:supervisor_prompt}, \ref{fig:database_prompt}, and \ref{fig:api_prompt}. For each prompt, we present a comprehensive specification detailing the agent’s responsibilities, operational workflow, prohibited actions, and exception handling procedures. This ensures clear role definitions and robust agent behavior throughout the system.

\end{document}